\documentclass[11pt]{article}

\usepackage[preprint]{acl}

\usepackage{times}
\usepackage{latexsym}
\usepackage[T1]{fontenc}
\usepackage[utf8]{inputenc}
\usepackage{microtype}
\usepackage{inconsolata}

\usepackage{graphicx}
\usepackage{amsmath,amsfonts,amssymb}
\usepackage{textcomp}
\usepackage{stfloats}
\usepackage{float}
\usepackage{url}
\usepackage{verbatim}
\usepackage[table]{xcolor}
\usepackage{tabularx, colortbl, booktabs, multirow, array}
\usepackage{makecell}
\usepackage{threeparttable}
\usepackage{pifont}
\usepackage{subcaption}
\newcolumntype{C}{>{\centering\arraybackslash}X}

\usepackage{algorithm}
\usepackage{algorithmic}

\usepackage{amsthm}

\usepackage{tcolorbox}
\tcbuselibrary{listings, breakable}
\usepackage{listings}
\newtcolorbox{promptbox}{
  breakable,
  colback=gray!3,
  colframe=gray!40,
  boxrule=0.4pt,
  arc=2pt,
  left=6pt,
  right=6pt,
  top=6pt,
  bottom=6pt
}

\definecolor{linkblue}{RGB}{0, 92, 175}

\hypersetup{
    colorlinks=true,
    linkcolor=linkblue,
    citecolor=linkblue,
    urlcolor=linkblue
}

\author{
Chaodong Tong\textsuperscript{1,2}
\quad
Qi Zhang\textsuperscript{3}
\quad
Nannan Sun\textsuperscript{1}
\quad
Lei Jiang\textsuperscript{1}
\quad
Yanbing Liu\textsuperscript{1,2}
\\[0.5em]
\textsuperscript{1}Institute of Information Engineering, Chinese Academy of Sciences, Beijing, China
\\
\textsuperscript{2}School of Cyber Security, University of Chinese Academy of Sciences, Beijing, China
\\
\textsuperscript{3}China Industrial Control Systems Cyber Emergency Response Team, Beijing, China
\\[0.5em]
\textsuperscript{1}\texttt{\{tongchaodong, sunnannan, jianglei, liuyanbing\}@iie.ac.cn}
\\
\textsuperscript{3}\texttt{bonniezhangqi@126.com}
}

\title{Knowledge Dependency Estimation for Reliable Question Answering}

\begin{document}

\maketitle

\begin{abstract}
Reliable question answering requires identifying not only whether an answer is correct, but also which available knowledge the prediction depends on.
In realistic LLM-based QA, this knowledge may come from context, retrieval, decomposition, or intermediate reasoning, forming a noisy and redundant candidate space rather than a clean gold evidence set.
We study \emph{knowledge dependency estimation}: estimating the sensitivity of a fixed black-box QA model to different candidate knowledge units.
The challenge is to obtain fine-grained dependency scores without exhaustive test-time perturbation while modeling redundancy, substitutability, and complementarity.
We propose \textbf{Knot}, a structured rank-aware knowledge dependency estimator.
Knot learns from subset-level counterfactual supervision, models subset sensitivity through coverage over latent dependency factors, and derives rank-aware unit scores to identify influential candidates.
Across multiple-choice and generative QA benchmarks, Knot outperforms all compared baselines in subset-sensitivity prediction and produces more faithful unit rankings than deployable baselines without extra QA-model calls; when used for practical risk screening, its dependency scores help flag error-prone QA predictions early.
\end{abstract}

\section{Introduction}
Large language models (LLMs) and modern question answering (QA) systems have achieved strong performance on knowledge-intensive and reasoning-oriented tasks, yet they still produce hallucinated facts, unstable reasoning, and overconfident predictions~\cite{lin2022teaching,kadavath2022language,manakul2023selfcheckgpt,farquhar2024detecting}. 
These failures raise a central question for trustworthy QA: beyond whether an answer appears correct, can we identify which knowledge the model's behavior actually depends on?

Prior work has studied related forms of reliability and explanation, but knowledge dependency in QA remains under-characterized.
Output-side uncertainty methods use predictive entropy, self-consistency, or semantic uncertainty to identify unreliable answers~\cite{manakul2023selfcheckgpt,farquhar2024detecting}.
Evidence-centered QA work extracts rationales or supporting evidence for answers~\cite{deyoung2020eraser}.
Perturbation-based and feature-attribution methods instead estimate which inputs or factors affect model behavior~\cite{li2016understanding,lundberg2017unified,covert2021explaining}.
These signals are valuable, but they do not directly estimate how much a QA model relies on the candidate knowledge provided at inference time, especially when that knowledge is noisy, heterogeneous, and redundant.
Such dependency information would enable inspection of influential evidence, assessment of evidence credibility, and detection of answers that may not be grounded in the provided candidates, motivating methods that explicitly estimate knowledge dependency.

We study knowledge dependency estimation under noisy candidate knowledge spaces, where a fixed QA model receives task context fragments, retrieved passages, decomposed information needs, or intermediate reasoning statements as candidate units.
The goal is to estimate the model's sensitivity to removing different candidates.
Because the signal is defined by changes in model behavior after candidate removal, it can help identify answer-driving knowledge, diagnose weak evidence dependence, and assess reliability.
The problem is challenging because exhaustive perturbation is costly, and isolated unit removal can be misleading when candidates overlap, substitute for one another, or jointly support an answer.
Similar difficulties arise in feature attribution and causal explanation, where correlated features and interaction effects complicate faithful importance estimation~\cite{tsang2018detecting,frye2020asymmetric,covert2020understanding,covert2021explaining}.

To address these challenges, we propose \textbf{Knot}, a structured rank-aware estimator for knowledge dependency sensitivity.
Knot learns from subset-level counterfactual supervision by observing behavioral changes after removing sampled candidate subsets.
It predicts subset sensitivity through latent factor coverage, where redundant candidates saturate shared factors and complementary evidence accumulates across factors.
Rank-aware supervision derived from subset labels further improves candidate-level dependency ordering.
Once trained, Knot scores candidate units in a single forward pass over the question and candidate space, avoiding repeated QA-model perturbation at inference time.
This formulation treats dependency as an operational property of a fixed QA model under a specified candidate space and perturbation protocol, complementing evidence extraction and rationale identification by focusing on how available knowledge shapes observed model behavior.

We evaluate Knot on multiple QA benchmarks covering both multiple-choice and generative settings.
The experiments assess subset-level sensitivity prediction, behavioral faithfulness of unit-level scores, component contributions, and downstream risk indication.
Knot outperforms all compared baselines in subset-sensitivity prediction, yields unit rankings that better match live perturbation effects than deployable baselines without additional QA-model calls, and helps identify error-prone QA predictions.

Our contributions are:
\begin{itemize}
    \item We introduce knowledge dependency estimation for noisy candidate knowledge spaces, framing it as behavioral sensitivity analysis for fixed QA models.
    \item We propose \textbf{Knot}, a structured rank-aware estimator that learns deployable unit-level dependency scores from subset-level counterfactual supervision.
    \item We demonstrate that dependency sensitivity can support both unit-level evidence-dependence analysis and practical risk screening across multiple-choice and generative QA settings.
\end{itemize}

\section{Problem Formulation}
\label{sec:problem}

We study knowledge dependency estimation for question answering (QA).
Given a question \(q\), a parameter-fixed black-box QA model \(f_\theta\), and a candidate knowledge space \(\mathcal{C}(q)\), the goal is to estimate how strongly the model's answer behavior depends on each candidate knowledge unit.
Such estimates can identify answer-driving knowledge units and support evidence-based reliability assessment.

\subsection{Candidate Knowledge Space}

For each question \(q\), we define a finite candidate knowledge space
\begin{equation}
\mathcal{C}(q)=\{k_1,k_2,\ldots,k_N\},
\label{eq:candidate_space}
\end{equation}
where each \(k_i\) may be a task-provided context fragment, retrieved passage, decomposed information need, or intermediate reasoning statement.
We do not assume that \(\mathcal{C}(q)\) is complete, minimal, or uniquely correct; it may contain redundant, partially relevant, or distracting units.

The QA model produces an answer conditioned on the question and candidate knowledge:
\begin{equation}
a
=
f_\theta(q,\mathcal{C}(q)).
\label{eq:qa_answer}
\end{equation}
Since \(f_\theta\) is treated as a black box, dependency is defined through
behavioral changes under candidate perturbations, rather than as unique evidence
or internal causal mechanisms.

\subsection{Knowledge Perturbation}

For a subset \(\mathcal{S}\subseteq\mathcal{C}(q)\), we define a perturbation operator
\begin{equation}
\phi(\mathcal{C}(q),\mathcal{S}),
\label{eq:perturbation_operator}
\end{equation}
which makes \(\mathcal{S}\) unavailable to the QA model, implemented in the main setting by removing it from the candidate context.
The perturbed answer is
\begin{equation}
a_{\mathcal{S}}
=
f_\theta\!\left(q,\phi(\mathcal{C}(q),\mathcal{S})\right).
\label{eq:perturbed_answer}
\end{equation}
The question is kept fixed, so the perturbation isolates the behavioral effect of changing the available candidate knowledge.

\subsection{Subset-level Knowledge Sensitivity}

We define the sensitivity of a removed subset \(\mathcal{S}\) as
\begin{equation}
S(q,\mathcal{S};\mathcal{C}(q))
=
\Delta(a,a_{\mathcal{S}};q),
\label{eq:subset_sensitivity}
\end{equation}
where \(\Delta(\cdot,\cdot;q)\in[0,1]\) is a task-dependent discrepancy function.
A larger value indicates a larger behavioral change after removing \(\mathcal{S}\).
The concrete construction of \(\Delta\) is described in Sec.~\ref{sec:method_supervision}.

\subsection{Unit-level Knowledge Sensitivity}

Our main target is the sensitivity of individual candidate units.
For each \(k_i\in\mathcal{C}(q)\), we define
\begin{equation}
S(q,k_i;\mathcal{C}(q))
=
S(q,\{k_i\};\mathcal{C}(q)),
\label{eq:unit_sensitivity}
\end{equation}
which measures the behavioral effect of making \(k_i\) unavailable while keeping the rest of the candidate context.

A direct leave-one-out procedure would require one additional QA-model call per candidate and must be repeated whenever \(\mathcal{C}(q)\) changes.
It also provides limited information about redundancy, substitutability, and complementarity among candidates.
We therefore learn a structured amortized estimator from subset-level counterfactual supervision:
\begin{equation}
\widehat{S}_{\Theta}(q,\mathcal{S};\mathcal{C}(q))
\approx
S(q,\mathcal{S};\mathcal{C}(q)),
\label{eq:subset_estimator}
\end{equation}
and obtain unit-level predictions by singleton evaluation:
\begin{equation}
\widehat{S}_{\Theta}(q,k_i;\mathcal{C}(q))
=
\widehat{S}_{\Theta}(q,\{k_i\};\mathcal{C}(q)).
\label{eq:unit_estimator}
\end{equation}

\begin{figure*}[t]
    \centering
    \includegraphics[width=\textwidth]{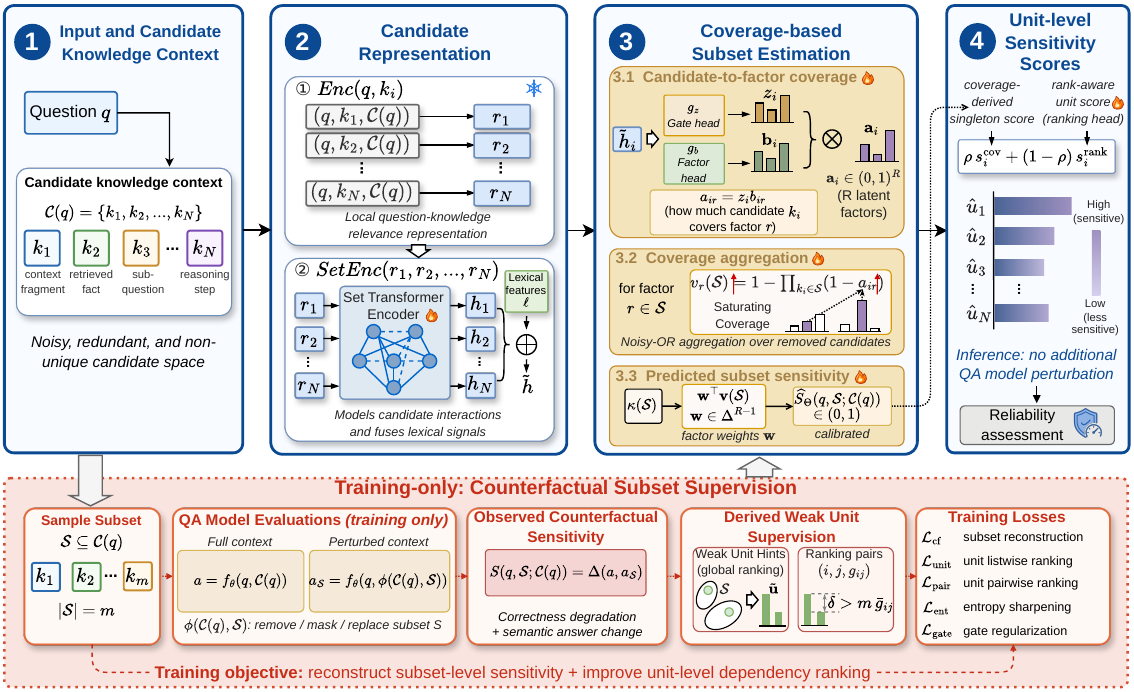}
    \caption{
    Overview of \textbf{Knot}.
    The model encodes a question and its noisy candidate knowledge space, predicts
    subset sensitivity through latent factor coverage, and derives unit-level
    dependency scores with rank-aware unit scoring.
    Training uses counterfactual subset perturbations, while inference requires no
    additional QA-model perturbation.
    }
    \label{fig:main_framework}
\end{figure*}

\section{Methodology}
\label{sec:method}
We propose \textbf{Knot}, a \emph{Structured Rank-aware Knowledge Dependency Estimator} that predicts how strongly a fixed QA model depends on each candidate unit \(k_i\in\mathcal{C}(q)\). Knot amortizes offline counterfactual perturbation signals into a deployable scorer, directly scoring candidates from (q) and \(\mathcal{C}(q)\) at inference time without additional QA-model calls. As shown in Fig.~\ref{fig:main_framework}, Knot estimates behavioral dependency rather than surface relevance: it encodes the candidate set, reconstructs subset-level sensitivity with a latent coverage estimator, and uses weak rank-aware supervision derived from subset labels to improve unit-level ordering. Thus, unit scores are learned in connection with subset behavior rather than from direct unit-level supervision.

\subsection{Counterfactual Supervision}
\label{sec:method_supervision}

For each question \(q\), the candidate space \(\mathcal{C}(q)\) consists of heterogeneous knowledge units, including task-context fragments, retrieved Wikipedia passages, generated subquestions, and generated reasoning-need statements.
Following perturbation-based and counterfactual attribution work~\cite{vig2020investigating,covert2021explaining,liu2025attribot}, we sample subsets \(\mathcal{S}\subseteq\mathcal{C}(q)\) and query the fixed QA model under the full candidate context and a perturbed context where \(\mathcal{S}\) is removed.
This yields the original answer \(a\) and the perturbed answer \(a_{\mathcal{S}}\).
The subset-level sensitivity label is
\begin{equation}
S(q,\mathcal{S};\mathcal{C}(q))
=
\Delta(a,a_{\mathcal{S}};q),
\label{eq:method_supervision_abstract}
\end{equation}
where \(\Delta\) measures the behavioral effect of removing \(\mathcal{S}\).

In our implementation, \(\Delta\) combines non-negative correctness degradation and semantic answer change:
\begin{equation}
\begin{aligned}
S(q,\mathcal{S};\mathcal{C}(q))
=
&\ \alpha
\left[
c(a;q)-c(a_{\mathcal{S}};q)
\right]_{+} \\
&+
(1-\alpha)\delta(a,a_{\mathcal{S}}),
\end{aligned}
\label{eq:method_supervision_label}
\end{equation}
where \(c(\cdot;q)\in[0,1]\) is a dataset-aware correctness score, \(\delta(\cdot,\cdot)\in[0,1]\) measures semantic answer change, \([x]_{+}=\max(x,0)\), and \(\alpha\in[0,1]\).
This construction turns the observed effect of subset perturbation into a bounded supervision signal for answer-level behavioral change under the chosen perturbation protocol.
Correctness is computed by normalized answer matching, task-specific answer-space matching, and an LLM judge when rule-based scoring is uncertain~\cite{gu2024survey}; answer change is computed by answer-space matching or semantic equivalence checking, following recent answer-equivalence evaluation for QA~\cite{li2024cfmatch}.

We also derive weak unit hints from the same subset labels.
For each candidate \(k_i\), let \(n_i\) be the number of sampled subsets containing \(k_i\), and define
\begin{equation}
\tilde{u}_i
=
\frac{1}{\max(1,n_i)}
\sum_{\mathcal{S}:k_i\in\mathcal{S}}
S(q,\mathcal{S};\mathcal{C}(q)).
\label{eq:weak_unit_hint}
\end{equation}
For candidate pairs with \(\tilde{u}_i>\tilde{u}_j\), we define the weak ranking gap
\begin{equation}
g_{ij}
=
\tilde{u}_i-\tilde{u}_j
>0.
\label{eq:weak_unit_pair}
\end{equation}
These pairs provide weak ordering constraints for the auxiliary ranking loss in Sec.~\ref{sec:method_training}: larger gaps encourage larger margins between the predicted unit scores.
Because they are computed from existing subset labels, these weak hints require no additional QA-model or LLM calls.
Details on candidate construction, subset sampling, and teacher scoring are provided in Appendix~\ref{app:data_supervision}.

\subsection{Candidate Representation}
\label{sec:method_representation}

For each candidate \(k_i\), we encode the question--candidate pair with a frozen text encoder:
\begin{equation}
r_i=\mathrm{Enc}(q,k_i),
\qquad
r_i\in\mathbb{R}^{d}.
\label{eq:method_local_encoding}
\end{equation}
The input contains only deployable information, including the question, candidate text, optional raw task context, and online-computable lexical features; it excludes reference answers, teacher outputs, supervision labels, and construction metadata.

Inspired by neural set modeling~\cite{lee2019set}, we use a set encoder with \(L=3\) layers to contextualize candidate representations and capture interactions among unordered candidates:
\begin{equation}
h_1,\ldots,h_N
=
\mathrm{SetEnc}(r_1,\ldots,r_N),
\qquad
h_i\in\mathbb{R}^{d}.
\label{eq:method_set_encoding}
\end{equation}
The full model further fuses lexical features:
\begin{equation}
\tilde{h}_i
=
h_i
+
\eta\,P_{\ell}(\ell_i),
\label{eq:method_lexical_fusion}
\end{equation}
where \(\ell_i\in\mathbb{R}^{d_\ell}\) is an online-available lexical feature vector, \(P_{\ell}\) projects it into \(\mathbb{R}^{d}\), and \(\eta\) is a learned fusion scale.
When lexical fusion is disabled, \(\tilde{h}_i=h_i\).
Implementation details are given in Appendix~\ref{app:kde_implementation}.

\subsection{Coverage-based Subset Estimation}
\label{sec:method_coverage}

Knot predicts subset sensitivity through latent factor coverage.
A gate estimates the context-dependent usefulness of each candidate:
\begin{equation}
z_i=\sigma(g_z(\tilde{h}_i)),
\qquad
z_i\in(0,1).
\label{eq:method_gate}
\end{equation}
A factor head predicts coverage over \(R=30\) latent dependency factors:
\begin{equation}
\mathbf{b}_i=\sigma(g_b(\tilde{h}_i)),
\qquad
\mathbf{b}_i\in(0,1)^R,
\label{eq:method_factor_raw}
\end{equation}
and the gated factor coverage is
\begin{equation}
\mathbf{a}_i=z_i\mathbf{b}_i,
\qquad
\mathbf{a}_i\in(0,1)^R .
\label{eq:method_gated_coverage}
\end{equation}

For a removed subset \(\mathcal{S}\), factor coverage is aggregated with a noisy-OR~\cite{pearl2014probabilistic}:
\begin{equation}
v_r(\mathcal{S})
=
1-
\prod_{k_i\in\mathcal{S}}
(1-a_{ir}),
\qquad
r=1,\ldots,R .
\label{eq:method_noisy_or}
\end{equation}
This aggregation saturates when multiple candidates cover the same factor, reducing over-counting of redundant units.
Let \(\mathbf{v}(\mathcal{S})=(v_1(\mathcal{S}),\ldots,v_R(\mathcal{S}))^\top\).
The subset coverage score is
\begin{equation}
\kappa(\mathcal{S})
=
\mathbf{w}^{\top}\mathbf{v}(\mathcal{S}),
\qquad
\mathbf{w}=\operatorname{softmax}(\boldsymbol{\beta}),
\label{eq:method_subset_coverage}
\end{equation}
where \(\boldsymbol{\beta}\in\mathbb{R}^{R}\), \(\mathbf{w}\in\Delta^{R-1}\), and \(\kappa(\mathcal{S})\in[0,1]\).

A monotonic calibration maps coverage to the sensitivity scale:
\begin{equation}
\widehat{S}_{\Theta}(q,\mathcal{S};\mathcal{C}(q))
=
\sigma\!\left(
\gamma\,
\operatorname{logit}
\left(
\operatorname{clip}_{\epsilon}(\kappa(\mathcal{S}))
\right)
\right),
\label{eq:method_subset_prediction}
\end{equation}
where \(\gamma=\operatorname{softplus}(\omega)>0\), and
\(\operatorname{clip}_{\epsilon}(x)=\min(1-\epsilon,\max(\epsilon,x))\) ensures numerical stability.

\subsection{Rank-aware Unit Scoring}
\label{sec:method_unit_scoring}

The coverage model gives a singleton score by setting \(\mathcal{S}=\{k_i\}\).
Since \(v_r(\{k_i\})=a_{ir}\), the coverage-derived unit score is
\begin{equation}
s^{\mathrm{cov}}_i
=
\sigma\!\left(
\gamma\,
\operatorname{logit}
\left(
\operatorname{clip}_{\epsilon}
\left(
\mathbf{w}^{\top}\mathbf{a}_i
\right)
\right)
\right).
\label{eq:method_coverage_unit_score}
\end{equation}

Subset reconstruction alone may leave individual candidate rankings under-constrained.
We therefore add a rank head:
\begin{equation}
s^{\mathrm{rank}}_i
=
\sigma(g_{\mathrm{rank}}(\tilde{h}_i)).
\label{eq:method_rank_head}
\end{equation}
The final unit dependency score is
\begin{equation}
\widehat{u}_i
=
\rho\,s^{\mathrm{cov}}_i
+
(1-\rho)\,s^{\mathrm{rank}}_i,
\qquad
\rho\in[0,1].
\label{eq:method_final_unit_score}
\end{equation}
The coverage term ties unit scores to the subset estimator, while the rank term improves candidate ordering under weak unit-ranking supervision.

\subsection{Training Objective}
\label{sec:method_training}

The primary loss reconstructs subset-level counterfactual sensitivity:
\begin{equation}
\mathcal{L}_{\mathrm{cf}}
=
\mathbb{E}_{q,\mathcal{S}}
\left[
\left(
\widehat{S}_{\Theta}(q,\mathcal{S};\mathcal{C}(q))
-
S(q,\mathcal{S};\mathcal{C}(q))
\right)^2
\right].
\label{eq:method_cf_loss}
\end{equation}

We use weak unit hints for auxiliary rank-aware supervision, combining listwise distribution matching and pairwise margin constraints inspired by learning-to-rank objectives~\cite{cao2007learning,burges2005learning}.
Let
\[
\tilde{\mathbf{u}}=(\tilde{u}_1,\ldots,\tilde{u}_N),
\qquad
\widehat{\mathbf{u}}=(\widehat{u}_1,\ldots,\widehat{u}_N).
\]
The listwise unit loss compares their within-question distributions:
\begin{equation}
\tilde{\boldsymbol{\pi}}
=
\operatorname{softmax}(\tilde{\mathbf{u}}/T),
\qquad
\widehat{\boldsymbol{\pi}}
=
\operatorname{softmax}(\widehat{\mathbf{u}}/T),
\label{eq:method_unit_distributions}
\end{equation}
\begin{equation}
\mathcal{L}_{\mathrm{unit}}
=
\mathrm{KL}
\left(
\tilde{\boldsymbol{\pi}}
\;\middle\|\;
\widehat{\boldsymbol{\pi}}
\right).
\label{eq:method_unit_listwise_loss}
\end{equation}

For valid weak ranking pairs \(\mathcal{P}=\{(i,j):\tilde{u}_i>\tilde{u}_j\}\), we use
\begin{equation}
\mathcal{L}_{\mathrm{pair}}
=
\frac{1}{|\mathcal{P}|}
\sum_{(i,j)\in\mathcal{P}}
\left[
m\,\bar{g}_{ij}
-
\widehat{u}_i
+
\widehat{u}_j
\right]_{+},
\label{eq:method_unit_pair_loss}
\end{equation}
where
\(\bar{g}_{ij}=\min(1,\max(10^{-4},\tilde{u}_i-\tilde{u}_j))\).
If \(\mathcal{P}\) is empty, this loss is set to zero.

We further add a light entropy-sharpening term over predicted unit scores and a gate regularizer to discourage uniformly high candidate usefulness.

The final objective is
\begin{equation}
\begin{aligned}
\mathcal{L}
=
&\ \mathcal{L}_{\mathrm{cf}}
+
\lambda_{\mathrm{unit}}\mathcal{L}_{\mathrm{unit}}
+
\lambda_{\mathrm{pair}}\mathcal{L}_{\mathrm{pair}}\\
&+
\lambda_{\mathrm{ent}}\mathcal{L}_{\mathrm{ent}}
+
\lambda_{\mathrm{gate}}\mathcal{L}_{\mathrm{gate}} .
\end{aligned}
\label{eq:method_final_loss}
\end{equation}
In the main implementation, we set \(\rho=0.5\), \(T=0.10\), \(m=0.10\), \(\lambda_{\mathrm{unit}}=0.20\), \(\lambda_{\mathrm{pair}}=0.20\), \(\lambda_{\mathrm{ent}}=0.005\), and \(\lambda_{\mathrm{gate}}=5\times10^{-5}\).
Full architectural and training details are provided in Appendix~\ref{app:kde_implementation}.

\section{Experiments}
\label{sec:experiments}
We evaluate \textbf{Knot} along four research questions.
RQ1 tests whether subset-level counterfactual supervision can be effectively learned by the amortized architecture.
RQ2 asks whether the induced unit scores align with live intervention effects on QA behavior.
RQ3 isolates the contribution of the main modeling components.
RQ4 evaluates whether knowledge dependency estimates provide useful signals for QA reliability.

\subsection{Experimental Setup}

\paragraph{Datasets.}
We evaluate on MMLU~\cite{hendrycks2021measuring}, MedQA~\cite{jin2021disease}, TruthfulQA~\cite{lin2022truthfulqa}, SQuAD~\cite{rajpurkar2016squad}, and HotpotQA~\cite{yang2018hotpotqa}.
The counterfactual supervision data used for training, validation, and the full-test deployable subset-estimation experiments is filtered to examples where the QA model answers correctly under the full candidate context.
For experiments involving costly comparison methods, we separately sample a 500-example subset from the original test split.
Dataset statistics are shown in Table~\ref{tab:dataset_statistics}; preprocessing details are in Appendix~\ref{app:datasets}.

\begin{table}[t]
\centering
\small
\setlength{\tabcolsep}{4pt}
\caption{
Dataset statistics.
MC denotes multiple-choice and GEN denotes generative short-answer QA.
}
\label{tab:dataset_statistics}
\begin{tabular}{llrrrr}
\toprule
Dataset & Task & Train & Dev & Test & Total \\
\midrule
MMLU & MC & 500 & 100 & 500 & 1100 \\
MedQA & MC & 500 & 100 & 500 & 1100 \\
TruthfulQA & MC & 311 & 62 & 311 & 684 \\
SQuAD & GEN & 500 & 100 & 500 & 1100 \\
HotpotQA & GEN & 500 & 100 & 500 & 1100 \\
\bottomrule
\end{tabular}
\end{table}

\paragraph{Candidate knowledge spaces.}
For each question, we construct \(\mathcal{C}(q)\) from task context, Wikipedia retrieval~\cite{lewis2020retrieval}, generated subquestions~\cite{ammann2025question},and generated reasoning needs~\cite{wei2022chain}.
As a result, candidates naturally include useful, redundant, incomplete, and distracting units.
Table~\ref{tab:cq_statistics} summarizes candidate statistics; retrieval and quality audits are in Appendix~\ref{app:candidate_construction} and~\ref{app:candidate_quality}.

\begin{table}[t]
\centering
\small
\setlength{\tabcolsep}{4pt}
\caption{
Candidate-space statistics.
Ctx., Retr., SubQ, and Reason. denote task-context, Wikipedia retrieval, generated subquestion, and generated reasoning-need.
Ex. denotes the number of questions in each split.
AnsCov. denotes answer coverage percentage.
}
\label{tab:cq_statistics}
\resizebox{\linewidth}{!}{%
\begin{tabular}{lrrrrrrrr}
\toprule
Split & Ex. & \(|\mathcal{C}|\) & Ctx. & Retr. & SubQ & Reason. & AnsCov. \\
\midrule
Train & 2311 & 10.7 & 1.3 & 5.0 & 2.7 & 1.8  & 52.4 \\
Dev   & 462  & 10.9 & 1.3 & 5.0 & 2.8 & 1.8  & 52.6 \\
Test  & 2311 & 10.9 & 1.3 & 5.0 & 2.8 & 1.8  & 51.8 \\
\bottomrule
\end{tabular}
}
\end{table}

\paragraph{Counterfactual supervision.}
For each sampled subset \(\mathcal{S}\subseteq\mathcal{C}(q)\), we compare teacher-model behavior under the full candidate space and under the perturbed space where \(\mathcal{S}\) is removed.
Following Section~\ref{sec:method_supervision}, the supervision target combines non-negative correctness drop and task-specific answer change, with \(\alpha=0.4\) in the main run.
All trainable methods use the same fixed supervision data.
Details of subset sampling, hybrid answer scoring, answer-equivalence testing, and supervision audits are provided in Appendices~\ref{app:subset_sampling} and~\ref{app:teacher_scoring}.

\paragraph{Compared methods.}
We compare \textbf{Knot} with deployable and expensive baselines.
Deployable baselines include BM25 relevance (\textbf{BM25})~\cite{robertson2009probabilistic}, size-only ridge regression (\textbf{Size})~\cite{hoerl1970ridge}, lexical-feature regressors using ridge, histogram gradient boosting~\cite{friedman2001greedy}, extremely randomized trees~\cite{geurts2006extremely}, and MLPs (\textbf{Lex-Ridge}, \textbf{Lex-HGB}, \textbf{Lex-ET}, \textbf{Lex-MLP}). Architectural ablations, including \textbf{w/o Latent}, are reported in RQ3.
On 500 sampled questions, we also evaluate usable but costly references: LLM-based ranking (\textbf{LLM-Rank}), leave-one-out perturbation (\textbf{LOO})~\cite{liu2025attribot}, and Monte Carlo Shapley (\textbf{MC-S4}, \textbf{MC-S16})~\cite{shapley1953value,covert2021explaining}.

\paragraph{Metrics.}
For subset-level prediction, we report MAE, RMSE, Pearson correlation, and Spearman correlation.
For unit-level faithfulness, we use intervention-based tests~\cite{deyoung2020eraser,jacovi2020towards,covert2021explaining,madsen2024self}: 
\emph{Drop@\(k\)} measures behavior change after removing the highest-scoring candidates, and
\emph{Suff.@\(k\)} measures preservation when only the highest-scoring candidates are kept.
We additionally report NDCG@\(k\)~\cite{jarvelin2002cumulated} for ranking-oriented unit evaluation.
For reliability assessment, we report risk concentration in the main text and AUROC/AUPRC in Appendix~\ref{app:reliability_details}, following answer correctness and confidence estimation work~\cite{li2024graph,yaldiz-etal-2025-design}.
Calls/Q denotes backend requests per question; figures may abbreviate this as Req./Q.

\begin{table}[t]
\centering
\scriptsize
\setlength{\tabcolsep}{4pt}
\caption{
Subset-level sensitivity prediction. Best results are bolded within each setting.
}
\label{tab:rq1_subset_prediction}
\resizebox{\linewidth}{!}{%
\begin{tabular}{clrrrr}
\toprule
Setting & Method & MAE$\downarrow$ & RMSE$\downarrow$ & Pear.$\uparrow$ & Spear.$\uparrow$ \\
\midrule
\multirow{7}{*}{\rotatebox[origin=c]{90}{Full}}
& Size        & 0.387 & 0.413 & 0.391 & 0.445 \\
& Lex-Ridge   & 0.329 & 0.382 & 0.523 & 0.541 \\
& Lex-HGB     & 0.277 & 0.355 & 0.610 & 0.613 \\
& Lex-ET      & 0.288 & 0.353 & 0.614 & 0.617 \\
& Lex-MLP     & 0.284 & 0.365 & 0.589 & 0.594 \\
& BM25        & 0.406 & 0.432 & 0.263 & 0.279 \\
& \textbf{Knot} & \textbf{0.211} & \textbf{0.313} & \textbf{0.716} & \textbf{0.699} \\
\midrule
\multirow{11}{*}{\rotatebox[origin=c]{90}{Sampled}}
& Size        & 0.388 & 0.412 & 0.389 & 0.444 \\
& Lex-Ridge   & 0.310 & 0.365 & 0.579 & 0.596 \\
& Lex-HGB     & 0.274 & 0.350 & 0.617 & 0.613 \\
& Lex-ET      & 0.279 & 0.345 & 0.633 & 0.624 \\
& Lex-MLP     & 0.277 & 0.364 & 0.583 & 0.591 \\
& BM25        & 0.399 & 0.426 & 0.288 & 0.325 \\
& LLM-Rank    & 0.399 & 0.562 & 0.408 & 0.479 \\
& LOO         & 0.394 & 0.560 & 0.236 & 0.262 \\
& MC-S4       & 0.273 & 0.390 & 0.559 & 0.569 \\
& MC-S16      & 0.272 & 0.383 & 0.570 & 0.577 \\
& \textbf{Knot} & \textbf{0.206} & \textbf{0.303} & \textbf{0.731} & \textbf{0.715} \\
\bottomrule
\end{tabular}
}
\end{table}

\subsection{RQ1: Subset Sensitivity Estimation}
\label{sec:rq1_subset}

Table~\ref{tab:rq1_subset_prediction} shows that \textbf{Knot} gives the strongest subset-level sensitivity estimates among the main deployable methods.
On the full retained test split, it reduces MAE from \(0.277\) for the strongest text baseline, Lex-HGB, to \(0.211\), while also achieving the best Pearson and Spearman correlations among deployable methods.
This indicates that Knot captures counterfactual dependency structure beyond subset size and lexical overlap.

\begin{figure}[t]
    \centering
    \includegraphics[width=\columnwidth]{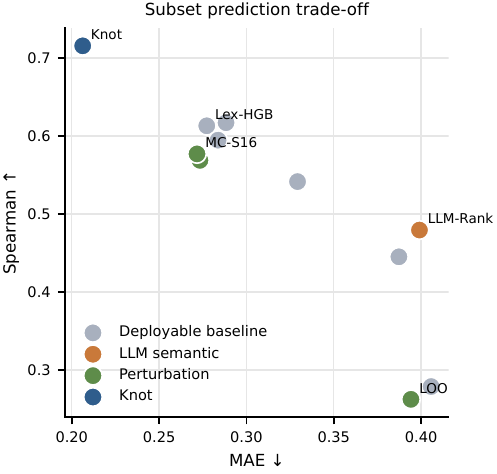}
    \caption{
    Subset sensitivity prediction trade-off.
    Lower MAE and higher Spearman are better.
    }
    \label{fig:rq1_tradeoff}
\end{figure}

\begin{figure}[t]
    \centering
    \includegraphics[width=\columnwidth]{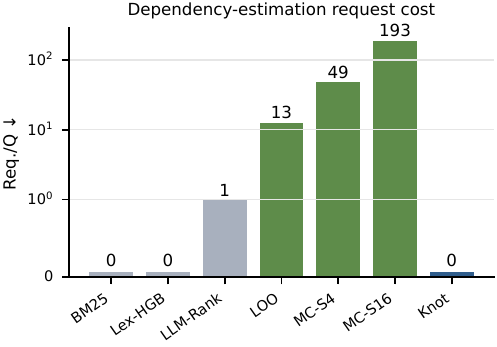}
    \caption{
    Inference-time backend cost.
    Costs assume \(N=12\) candidates.
    }
    \label{fig:rq1_cost_bar}
\end{figure}

The same pattern holds on the original-test sampled set.
Even without test-time QA perturbations, Knot outperforms LLM-Rank, LOO, and MC-Shapley across all subset-prediction metrics.
Fig.~\ref{fig:rq1_tradeoff} visualizes this result: Knot occupies the favorable low-error, high-correlation region.
Fig.~\ref{fig:rq1_cost_bar} highlights the main practical implication.
Perturbation-based references require repeated backend calls, whereas Knot amortizes those signals into a one-pass dependency estimator.

\subsection{RQ2: Behavioral Unit Faithfulness}
\label{sec:rq2_faithfulness}

RQ2 asks whether the estimated unit scores identify candidates whose removal or
retention produces larger observable changes in QA behavior.
For each method, we rank candidates, either remove the top-ranked units or keep
only them, and measure the resulting behavioral change.
Higher Drop@\(k\) and Suff.@\(k\) indicate larger removal effects and stronger
sufficiency, respectively.

\begin{figure}[t]
    \centering
    \includegraphics[width=\columnwidth]{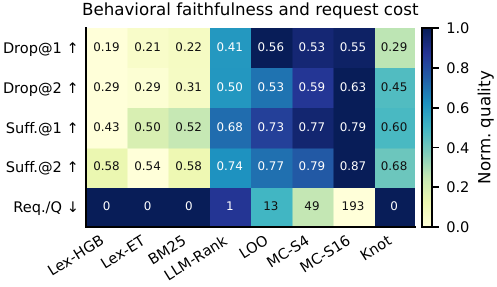}
    \caption{
    Behavioral faithfulness vs. request cost.
    Higher is better except for Req./Q, the number of backend requests per question.
    }
    \label{fig:rq2_behavior_cost}
\end{figure}

Fig.~\ref{fig:rq2_behavior_cost} shows the trade-off between intervention
faithfulness and evaluation cost.
LOO and MC-Shapley directly estimate counterfactual effects with test-time
QA-model calls; with enough subset samples, MC-Shapley can be viewed as a
black-box upper reference for intervention-based unit scoring.
LLM-Rank reduces the cost to a single ranking prompt, but it depends on an
external judge and can be sensitive to prompt design and candidate-list length.
Deployable estimators avoid test-time QA or LLM calls altogether.
Within this zero-call setting, Knot yields the strongest behavioral alignment,
suggesting that subset-level supervision can be amortized into meaningful
unit-level dependency scores without additional inference-time calls.

\subsection{RQ3: Component Ablation}
\label{sec:rq3_ablation}

RQ3 isolates the contribution of Knot's main components.
We ablate the rank-aware unit objective, lexical fusion, latent coverage, and the unit gate under the same training data and evaluation protocol.
Architectural hyperparameter ablations and loss-function variants are reported in Appendices~\ref{app:arch_hyperparam_ablation} and~\ref{app:optimization_variants}.

\begin{table}[t]
\centering
\small
\setlength{\tabcolsep}{2.6pt}
\caption{
Architectural ablations.
All metrics are measured on the held-out test split.
}
\label{tab:rq3_architectural_ablation}
\begin{tabular}{lrrrr}
\toprule
Variant & MAE$\downarrow$ & RMSE$\downarrow$ & Pear.$\uparrow$ & Spear.$\uparrow$ \\
\midrule
Knot & \textbf{0.211} & \textbf{0.313} & \textbf{0.716} & \textbf{0.699} \\
w/o Rank & 0.212 & 0.323 & 0.703 & 0.685 \\
w/o Lex. & 0.242 & 0.343 & 0.655 & 0.637 \\
w/o Latent & 0.244 & 0.358 & 0.615 & 0.604 \\
w/o Gate & 0.224 & 0.315 & 0.710 & 0.694 \\
\bottomrule
\end{tabular}
\end{table}

Table~\ref{tab:rq3_architectural_ablation} shows that lexical fusion and latent coverage are the most influential components.
Removing lexical fusion increases MAE from \(0.211\) to \(0.242\) and reduces Spearman from \(0.699\) to \(0.637\), while removing latent coverage yields the largest ranking degradation, with Spearman dropping to \(0.604\).
These results suggest that lexical signals complement neural candidate representations and that shared latent factors help capture redundancy and substitutability.
The rank-aware objective and unit gate also provide consistent gains, though their effects are smaller.
Overall, Knot achieves the best results across all four metrics, with the additive \textbf{w/o Latent} variant further discussed in Appendix~\ref{app:factor_vs_additive}.

\subsection{RQ4: Dependency-Based Reliability Assessment}
\label{sec:rq4_reliability}

RQ4 tests whether dependency estimates are useful for reliability assessment beyond attribution.
We score full-context QA outputs on the sampled test questions and ask whether dependency signals can rank answer errors.
For each method, we use its unit scores to select top-ranked knowledge units and derive behavior-audited risk scores from removal drop and sufficiency; details are given in Appendix~\ref{app:reliability_details}.
This treats dependency as complementary to output-level uncertainty: answers that are highly sensitive to a few selected units, yet not sufficiently supported by them alone, may be more vulnerable.

\begin{figure}[t]
    \centering
    \includegraphics[width=\columnwidth]{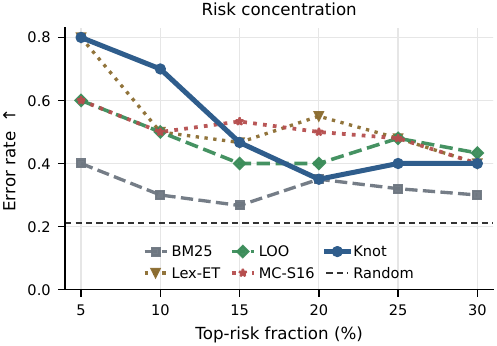}
    \caption{
    Dependency-based risk concentration using top-2 selected units.
    }
    \label{fig:rq4_risk_concentration}
\end{figure}

Fig.~\ref{fig:rq4_risk_concentration} shows that dependency-based risk scores concentrate answer errors above the random baseline in the low-budget review regime.
At a \(10\%\) review budget, the behavior-audited score based on Knot's top-2 predicted units flags answers with a \(0.700\) error rate, yielding a \(3.33\times\) lift over the overall error rate.
Risk@2 is an evaluation-time audit score, while the unit ranking used to select audited units is produced by Knot without live perturbation.
Thus, amortized dependency estimation can help prioritize potentially unreliable answers for verification.

\section{Related Work}
\label{sec:related_work} 
\paragraph{Explanation and evidence in QA.} Explainable QA and LLM reasoning work studies rationale extraction, supporting-evidence identification, retrieval-augmented evidence selection, source attribution, and reasoning decomposition~\cite{deyoung2020eraser,lewis2020retrieval,nematov2025source,wei2022chain}. These methods typically produce evidence, source links, or reasoning traces, often asking whether an answer can be attributed to retrieved documents. Knot instead estimates behavioral dependency scores for noisy, non-unique candidate knowledge units, asking how each unit shapes model behavior.

\paragraph{Perturbation-based and structured attribution.} Perturbation-based explanation estimates importance from prediction changes under input removal, editing, or intervention~\cite{li2016understanding,lundberg2017unified,covert2021explaining}. Recent QA attribution benchmarks show that identifying evidence that actually supports a model's answer remains difficult in realistic QA settings~\cite{hu2025can}. These methods motivate our counterfactual supervision, but direct LOO or Shapley-style estimation requires repeated model calls and is unstable under redundant or substitutable evidence. Knot learns a reusable estimator from subset-level labels and uses latent coverage to model redundancy and complementarity. 

\paragraph{Reliability, uncertainty, and amortization.} Uncertainty and hallucination detection estimate reliability from confidence, sampling consistency, or semantic uncertainty~\cite{manakul2023selfcheckgpt,kuhn2023semantic,farquhar2024detecting}. Our signal is complementary: it measures how answer behavior depends on available knowledge. Knot is also related to amortized inference and interaction-aware attribution, which approximate expensive subset-contribution procedures with reusable models~\cite{gershman2014amortized,tsang2018detecting,covert2021explaining}. A more comprehensive review is provided in Appendix~\ref{app:extended_related_work}.

\section{Conclusion}
We presented \textbf{Knot}, a structured framework for estimating how QA models depend on noisy, heterogeneous candidate knowledge.
Knot uses subset-level counterfactual supervision to learn behavior-aware unit scores, allowing dependency estimation to be performed without repeated QA-model perturbations at inference time.

Our results show that Knot improves subset-sensitivity prediction, produces more behaviorally aligned unit rankings among deployable methods, and provides useful signals for reliability assessment.
These findings suggest that knowledge dependency is a valuable abstraction for moving beyond surface relevance toward behavior-aware evaluation of knowledge-grounded QA.
Future work can extend this direction to richer candidate spaces, longer-form reasoning, and reliability-aware retrieval or verification pipelines.

\section*{Limitations}
Our study estimates dependency with respect to the candidate knowledge space available to the QA model.
Thus, the scores naturally reflect the coverage and granularity of candidate construction.
Future work can further improve candidate coverage and granularity, and study how dependency signals can guide retrieval or candidate refinement.

The counterfactual supervision also depends on the evaluated QA model and answer-equivalence scoring.
We use task-aware matching, semantic checks, and hybrid correctness scoring to reduce spurious differences, but richer answer types may require more fine-grained semantic evaluation.

Finally, Knot learns from sampled subset perturbations.
This makes inference efficient, while leaving room to better cover rare higher-order interactions.
Although the framework is model-agnostic, our experiments are limited by computational resources to controlled model and dataset settings.
Broader evaluations across more QA backends, retrieval settings, and adaptive sampling strategies would further clarify its deployment behavior.


\bibliography{references}

@inproceedings{ammann2025question,
    title = "Question Decomposition for Retrieval-Augmented Generation",
    author = "Ammann, Paul J. L.  and
      Golde, Jonas  and
      Akbik, Alan",
    booktitle = "Proceedings of the 63rd Annual Meeting of the Association for Computational Linguistics (Volume 4: Student Research Workshop)",
    year = "2025",
    pages = "497--507",

}

@inproceedings{burges2005learning,
  title={Learning to rank using gradient descent},
  author={Burges, Chris and Shaked, Tal and Renshaw, Erin and Lazier, Ari and Deeds, Matt and Hamilton, Nicole and Hullender, Greg},
  booktitle={Proceedings of the 22nd international conference on Machine learning},
  pages={89--96},
  year={2005}
}

@inproceedings{cao2007learning,
  title={Learning to rank: from pairwise approach to listwise approach},
  author={Cao, Zhe and Qin, Tao and Liu, Tie-Yan and Tsai, Ming-Feng and Li, Hang},
  booktitle={Proceedings of the 24th international conference on Machine learning},
  pages={129--136},
  year={2007}
}

@article{covert2020understanding,
  title={Understanding global feature contributions with additive importance measures},
  author={Covert, Ian and Lundberg, Scott M and Lee, Su-In},
  journal={Advances in neural information processing systems},
  volume={33},
  pages={17212--17223},
  year={2020}
}

@article{covert2021explaining,
  title={Explaining by removing: A unified framework for model explanation},
  author={Covert, Ian and Lundberg, Scott and Lee, Su-In},
  journal={Journal of Machine Learning Research},
  volume={22},
  number={209},
  pages={1--90},
  year={2021}
}

@inproceedings{deyoung2020eraser,
  title={ERASER: A benchmark to evaluate rationalized NLP models},
  author={DeYoung, Jay and Jain, Sarthak and Rajani, Nazneen Fatema and Lehman, Eric and Xiong, Caiming and Socher, Richard and Wallace, Byron C},
  booktitle={Proceedings of the 58th annual meeting of the association for computational linguistics},
  pages={4443--4458},
  year={2020}
}

@article{farquhar2024detecting,
  title={Detecting hallucinations in large language models using semantic entropy},
  author={Farquhar, Sebastian and Kossen, Jannik and Kuhn, Lorenz and Gal, Yarin},
  journal={Nature},
  volume={630},
  number={8017},
  pages={625--630},
  year={2024},
  publisher={Nature Publishing Group UK London}
}

@article{friedman2001greedy,
  title={Greedy function approximation: a gradient boosting machine},
  author={Friedman, Jerome H},
  journal={Annals of statistics},
  pages={1189--1232},
  year={2001},
  publisher={JSTOR}
}

@article{frye2020asymmetric,
  title={Asymmetric shapley values: incorporating causal knowledge into model-agnostic explainability},
  author={Frye, Christopher and Rowat, Colin and Feige, Ilya},
  journal={Advances in neural information processing systems},
  volume={33},
  pages={1229--1239},
  year={2020}
}

@inproceedings{gershman2014amortized,
  title={Amortized inference in probabilistic reasoning},
  author={Gershman, Samuel J. and Goodman, Noah D.},
  booktitle={Proceedings of the 36th Annual Conference of the Cognitive Science Society},
  pages={517--522},
  year={2014}
}

@article{geurts2006extremely,
  title={Extremely randomized trees},
  author={Geurts, Pierre and Ernst, Damien and Wehenkel, Louis},
  journal={Machine learning},
  volume={63},
  number={1},
  pages={3--42},
  year={2006},
  publisher={Springer}
}

@article{gu2024survey,
  title={A Survey on LLM-as-a-Judge},
  author={Gu, Jiawei and Jiang, Xuhui and Shi, Zhichao and Tan, Hexiang and Zhai, Xuehao and Xu, Chengjin and Li, Wei and Shen, Yinghan and Ma, Shengjie and Liu, Honghao and others},
  journal={arXiv preprint arXiv:2411.15594},
  year={2024}
}

@inproceedings{
hendrycks2021measuring,
title={Measuring Massive Multitask Language Understanding},
author={Dan Hendrycks and Collin Burns and Steven Basart and Andy Zou and Mantas Mazeika and Dawn Song and Jacob Steinhardt},
booktitle={International Conference on Learning Representations},
year={2021},
}

@article{hoerl1970ridge,
  title={Ridge regression: Biased estimation for nonorthogonal problems},
  author={Hoerl, Arthur E and Kennard, Robert W},
  journal={Technometrics},
  volume={12},
  number={1},
  pages={55--67},
  year={1970},
  publisher={Taylor \& Francis}
}

@inproceedings{hu2025can,
  title={Can LLMs Evaluate Complex Attribution in QA? Automatic Benchmarking using Knowledge Graphs},
  author={Hu, Nan and Chen, Jiaoyan and Wu, Yike and Qi, Guilin and Wang, Hongru and Bi, Sheng and Chen, Yongrui and Wu, Tongtong and Pan, Jeff Z},
  booktitle={Proceedings of the 63rd Annual Meeting of the Association for Computational Linguistics (Volume 1: Long Papers)},
  pages={17096--17118},
  year={2025}
}

@article{izacard2023atlas,
  title={Atlas: Few-shot learning with retrieval augmented language models},
  author={Izacard, Gautier and Lewis, Patrick and Lomeli, Maria and Hosseini, Lucas and Petroni, Fabio and Schick, Timo and Dwivedi-Yu, Jane and Joulin, Armand and Riedel, Sebastian and Grave, Edouard},
  journal={Journal of Machine Learning Research},
  volume={24},
  number={251},
  pages={1--43},
  year={2023}
}

@inproceedings{jacovi2020towards,
  title={Towards faithfully interpretable NLP systems: How should we define and evaluate faithfulness?},
  author={Jacovi, Alon and Goldberg, Yoav},
  booktitle={Proceedings of the 58th annual meeting of the association for computational linguistics},
  pages={4198--4205},
  year={2020}
}

@article{jin2021disease,
  title={What disease does this patient have? a large-scale open domain question answering dataset from medical exams},
  author={Jin, Di and Pan, Eileen and Oufattole, Nassim and Weng, Wei-Hung and Fang, Hanyi and Szolovits, Peter},
  journal={Applied Sciences},
  volume={11},
  number={14},
  pages={6421},
  year={2021},
}

@article{kadavath2022language,
  title={Language models (mostly) know what they know},
  author={Kadavath, Saurav and Conerly, Tom and Askell, Amanda and Henighan, Tom and Drain, Dawn and Perez, Ethan and Schiefer, Nicholas and Hatfield-Dodds, Zac and DasSarma, Nova and Tran-Johnson, Eli and others},
  journal={arXiv preprint arXiv:2207.05221},
  year={2022}
}

@inproceedings{kaushik2020learning,
  title={Learning The Difference That Makes A Difference With Counterfactually-Augmented Data},
  author={Kaushik, Divyansh and Hovy, Eduard and Lipton, Zachary C},
  booktitle={International Conference on Learning Representations (ICLR)},
  year={2020}
}

@inproceedings{
kuhn2023semantic,
title={Semantic Uncertainty: Linguistic Invariances for Uncertainty Estimation in Natural Language Generation},
author={Lorenz Kuhn and Yarin Gal and Sebastian Farquhar},
booktitle={The Eleventh International Conference on Learning Representations },
year={2023}
}

@article{kossen2024semantic,
  title={Semantic entropy probes: Robust and cheap hallucination detection in llms},
  author={Kossen, Jannik and Han, Jiatong and Razzak, Muhammed and Schut, Lisa and Malik, Shreshth and Gal, Yarin},
  journal={arXiv preprint arXiv:2406.15927},
  year={2024}
}

@inproceedings{lee2019set,
  title={Set transformer: A framework for attention-based permutation-invariant neural networks},
  author={Lee, Juho and Lee, Yoonho and Kim, Jungtaek and Kosiorek, Adam and Choi, Seungjin and Teh, Yee Whye},
  booktitle={International conference on machine learning},
  pages={3744--3753},
  year={2019},
  organization={PMLR}
}

@article{lewis2020retrieval,
  title={Retrieval-augmented generation for knowledge-intensive nlp tasks},
  author={Lewis, Patrick and Perez, Ethan and Piktus, Aleksandra and Petroni, Fabio and Karpukhin, Vladimir and Goyal, Naman and K{\"u}ttler, Heinrich and Lewis, Mike and Yih, Wen-tau and Rockt{\"a}schel, Tim and others},
  journal={Advances in neural information processing systems},
  volume={33},
  pages={9459--9474},
  year={2020}
}

@article{li2016understanding,
  title={Understanding neural networks through representation erasure},
  author={Li, Jiwei and Monroe, Will and Jurafsky, Dan},
  journal={arXiv preprint arXiv:1612.08220},
  year={2016}
}

@article{li2024cfmatch,
  title={CFMatch: Aligning automated answer equivalence evaluation with expert judgments for open-domain question answering},
  author={Li, Zongxia and Mondal, Ishani and Liang, Yijun and Nghiem, Huy and Boyd-Graber, Jordan},
  journal={arXiv preprint arXiv:2401.13170},
  year={2024}
}

@article{li2024graph,
  title={Graph-based confidence calibration for large language models},
  author={Li, Yukun and Wang, Sijia and Huang, Lifu and Liu, Li-Ping},
  journal={arXiv preprint arXiv:2411.02454},
  year={2024}
}

@article{
lin2022teaching,
title={Teaching Models to Express Their Uncertainty in Words},
author={Stephanie Lin and Jacob Hilton and Owain Evans},
journal={Transactions on Machine Learning Research},
issn={2835-8856},
year={2022}
}

@inproceedings{lin2022truthfulqa,
  title={Truthfulqa: Measuring how models mimic human falsehoods},
  author={Lin, Stephanie and Hilton, Jacob and Evans, Owain},
  booktitle={Proceedings of the 60th annual meeting of the association for computational linguistics (volume 1: long papers)},
  pages={3214--3252},
  year={2022}
}

@inproceedings{
liu2025attribot,
title={AttriBoT: A Bag of Tricks for Efficiently Approximating Leave-One-Out Context Attribution},
author={Fengyuan Liu and Nikhil Kandpal and Colin Raffel},
booktitle={The Thirteenth International Conference on Learning Representations},
year={2025},
}

@article{lundberg2017unified,
  title={A unified approach to interpreting model predictions},
  author={Lundberg, Scott M and Lee, Su-In},
  journal={Advances in neural information processing systems},
  volume={30},
  year={2017}
}

@inproceedings{madsen2024self,
  title={Are self-explanations from Large Language Models faithful?},
  author={Madsen, Andreas and Chandar, Sarath and Reddy, Siva},
  booktitle={Findings of the Association for Computational Linguistics: ACL 2024},
  pages={295--337},
  year={2024}
}

@inproceedings{manakul2023selfcheckgpt,
  title={Selfcheckgpt: Zero-resource black-box hallucination detection for generative large language models},
  author={Manakul, Potsawee and Liusie, Adian and Gales, Mark},
  booktitle={Proceedings of the 2023 conference on empirical methods in natural language processing},
  pages={9004--9017},
  year={2023}
}

@inproceedings{marino2018iterative,
  title={Iterative amortized inference},
  author={Marino, Joe and Yue, Yisong and Mandt, Stephan},
  booktitle={International Conference on Machine Learning},
  pages={3403--3412},
  year={2018},
  organization={PMLR}
}

@article{meng2022locating,
  title={Locating and editing factual associations in gpt},
  author={Meng, Kevin and Bau, David and Andonian, Alex and Belinkov, Yonatan},
  journal={Advances in neural information processing systems},
  volume={35},
  pages={17359--17372},
  year={2022}
}

@inproceedings{nematov2025source,
  title={Source attribution in retrieval-augmented generation},
  author={Nematov, Ikhtiyor and Kalai, Tarik and Kuzmenko, Elizaveta and Fugagnoli, Gabriele and Sacharidis, Dimitris and Hose, Katja and Sagi, Tomer},
  booktitle={Joint European Conference on Machine Learning and Knowledge Discovery in Databases},
  pages={317--332},
  year={2025},
  organization={Springer}
}

@book{pearl2014probabilistic,
  title={Probabilistic reasoning in intelligent systems: networks of plausible inference},
  author={Pearl, Judea},
  year={2014},
  publisher={Elsevier}
}

@inproceedings{rajpurkar2016squad,
  title={Squad: 100,000+ questions for machine comprehension of text},
  author={Rajpurkar, Pranav and Zhang, Jian and Lopyrev, Konstantin and Liang, Percy},
  booktitle={Proceedings of the 2016 conference on empirical methods in natural language processing},
  pages={2383--2392},
  year={2016}
}

@book{robertson2009probabilistic,
  title={The probabilistic relevance framework: BM25 and beyond},
  author={Robertson, Stephen and Zaragoza, Hugo},
  volume={4},
  year={2009},
  publisher={Now Publishers Inc}
}

@incollection{shapley1953value,
  title = {A Value for n-Person Games},
  author = {Shapley, Lloyd S},
  booktitle = {Contributions to the Theory of Games II},
  editor = {Kuhn, Harold W. and Tucker, Albert W.},
  pages = {307--317},
  year = {1953},
  publisher = {Princeton University Press},
  address = {Princeton}
}

@inproceedings{
tsang2018detecting,
title={Detecting Statistical Interactions from Neural Network Weights},
author={Michael Tsang and Dehua Cheng and Yan Liu},
booktitle={International Conference on Learning Representations},
year={2018},
}

@article{vig2020investigating,
  title={Investigating gender bias in language models using causal mediation analysis},
  author={Vig, Jesse and Gehrmann, Sebastian and Belinkov, Yonatan and Qian, Sharon and Nevo, Daniel and Singer, Yaron and Shieber, Stuart},
  journal={Advances in neural information processing systems},
  volume={33},
  pages={12388--12401},
  year={2020}
}

@inproceedings{
wang2023selfconsistency,
title={Self-Consistency Improves Chain of Thought Reasoning in Language Models},
author={Xuezhi Wang and Jason Wei and Dale Schuurmans and Quoc V Le and Ed H. Chi and Sharan Narang and Aakanksha Chowdhery and Denny Zhou},
booktitle={The Eleventh International Conference on Learning Representations },
year={2023},
}

@article{wei2022chain,
  title={Chain-of-thought prompting elicits reasoning in large language models},
  author={Wei, Jason and Wang, Xuezhi and Schuurmans, Dale and Bosma, Maarten and Xia, Fei and Chi, Ed and Le, Quoc V and Zhou, Denny and others},
  journal={Advances in neural information processing systems},
  volume={35},
  pages={24824--24837},
  year={2022}
}

@article{wu2024gendec,
  title={Gendec: A robust generative question-decomposition method for multi-hop reasoning},
  author={Wu, Jian and Yang, Linyi and Ji, Yuliang and Huang, Wenhao and Karlsson, B{\"o}rje F and Okumura, Manabu},
  journal={arXiv preprint arXiv:2402.11166},
  year={2024}
}

@inproceedings{yaldiz-etal-2025-design,
  title={Do not design, learn: A trainable scoring function for uncertainty estimation in generative LLMs},
  author={Yaldiz, Duygu Nur and Bakman, Yavuz Faruk and Buyukates, Baturalp and Tao, Chenyang and Ramakrishna, Anil and Dimitriadis, Dimitrios and Zhao, Jieyu and Avestimehr, Salman},
  booktitle={Findings of the Association for Computational Linguistics: NAACL 2025},
  pages={691--713},
  year={2025}
}

@inproceedings{yang2018hotpotqa,
  title={HotpotQA: A dataset for diverse, explainable multi-hop question answering},
  author={Yang, Zhilin and Qi, Peng and Zhang, Saizheng and Bengio, Yoshua and Cohen, William and Salakhutdinov, Ruslan and Manning, Christopher D},
  booktitle={Proceedings of the 2018 conference on empirical methods in natural language processing},
  pages={2369--2380},
  year={2018}
}

@article{jarvelin2002cumulated,
  title={Cumulated gain-based evaluation of IR techniques},
  author={J{\"a}rvelin, Kalervo and Kek{\"a}l{\"a}inen, Jaana},
  journal={ACM Transactions on Information Systems (TOIS)},
  volume={20},
  number={4},
  pages={422--446},
  year={2002},
  publisher={ACM New York, NY, USA}
}

@inproceedings{tong2026semantic,
  title={Semantic reformulation entropy for robust hallucination detection in qa tasks},
  author={Tong, Chaodong and Zhang, Qi and Jiang, Lei and Liu, Yanbing and Sun, Nannan and Li, Wei},
  booktitle={ICASSP 2026-2026 IEEE International Conference on Acoustics, Speech and Signal Processing (ICASSP)},
  pages={3381--3385},
  year={2026},
  organization={IEEE}
}

\clearpage
\appendix
\section{Data Construction and Counterfactual Supervision}
\label{app:data_supervision}

This appendix provides the data and supervision details needed to interpret the experiments and reproduce \textbf{Knot}.
The main text defines the knowledge-dependency estimation problem and reports the core results; here we specify the implementation choices that shape the supervision signal, including how the heterogeneous candidate space \(\mathcal{C}(q)\) is constructed, how counterfactual subsets are sampled, and how teacher behavior is converted into sensitivity labels.

\subsection{Dataset Sampling and Unified Schema}
\label{app:datasets}

We evaluate on MMLU, MedQA, TruthfulQA, SQuAD, and HotpotQA.
For each dataset, we sample up to 500/100/500 examples for train/dev/test; TruthfulQA contains fewer usable examples under the selected multiple-choice setup, so all available examples are retained.
All examples are converted into a common schema containing the dataset name, split, question, reference answer, task type, optional task context, and an answer space when applicable.
For multiple-choice datasets, option labels and option texts are canonicalized into the same answer space, since teacher answers may use either an option letter or an option string.
The resulting dataset statistics are reported in Table~\ref{tab:dataset_statistics} in the main paper.

\subsection{Candidate Knowledge Space Construction}
\label{app:candidate_construction}

For each question \(q\), we construct a noisy candidate knowledge space \(\mathcal{C}(q)\) from multiple sources that resemble the knowledge acquisition channels used by modern LLM-based QA systems, including task-provided context, external retrieval, generated subquestions, and generated reasoning needs where applicable.
The resulting candidate pool may contain useful, redundant, incomplete, or distracting units.
By making such knowledge candidates discrete and explicit, we can define knowledge dependency behaviorally through the QA model's response to perturbations over \(\mathcal{C}(q)\).

\paragraph{Task-context candidates.}
When the original dataset provides context, we split it into sentence-level or short-passage units and keep compact fragments that are relevant to the question.
This source is mainly active for SQuAD and HotpotQA.
For multiple-choice tasks, answer options remain part of the question definition and are not treated as removable knowledge units, because deleting an option changes the task rather than perturbing evidence.

\paragraph{Wikipedia retrieval candidates.}
External candidates are retrieved from a txtai\footnote{\url{https://neuml.github.io/txtai/}} index over the Hugging Face \texttt{wikimedia/wikipedia} corpus, version \texttt{20231101.en}.
We perform retrieval without access to the reference answer, so that the construction process better approximates a realistic QA setting in which potentially relevant knowledge must be gathered without answer leakage.
Queries are formed from information available at candidate-construction time, including question keywords, task-context anchors, title or entity cues, option texts for multiple-choice tasks, and dataset-specific query cues.
The reference answer is not used to form retrieval queries or rank retrieved passages.
Retrieved passages are hydrated from the passage map, lightly reranked using lexical overlap, title-anchor matches, and dataset-aware profile matches, and then length-controlled before insertion into \(\mathcal{C}(q)\).
Table~\ref{tab:wiki_retrieval_audit} reports a retrieval audit.

\begin{table}[t]
\centering
\small
\caption{
Sampled Wikipedia retrieval audit for external candidate construction.
\(N\) denotes the number of audited questions.
Q-Overlap is the best lexical overlap between the question and any retrieved passage.
Title Hit measures whether retrieved Wikipedia titles match extracted title or entity cues.
Profile Hit measures whether the retrieved pool matches dataset-aware query profiles.
}
\label{tab:wiki_retrieval_audit}
\resizebox{\linewidth}{!}{%
\begin{tabular}{lrrrr}
\toprule
Dataset & \(N\) & Q-Overlap & Title Hit & Profile Hit \\
\midrule
MMLU & 100 & 26.7 & 74.0 & 94.0 \\
MedQA & 100 & 10.1 & 85.0 & 94.0 \\
TruthfulQA & 100 & 37.0 & 82.0 & 92.0 \\
SQuAD & 100 & 34.1 & 79.0 & 99.0 \\
HotpotQA & 100 & 29.8 & 89.0 & 100.0 \\
Overall & 500 & 27.5 & 81.8 & 95.8 \\
\bottomrule
\end{tabular}
}
\end{table}

\paragraph{Generated subquestions and reasoning needs.}
We use a local LLM\footnote{
We use \texttt{Qwen/Qwen3.5-35B-A3B}. 
For a controlled trade-off among resource cost, throughput, and model capability, this is the same backend used as the QA model and teacher model in our experiments.
}
to generate answer-agnostic subquestions that describe missing information needs for solving the original question.
We also generate concise reasoning-need statements or bridge facts.
These candidates are inspired by question decomposition and chain-of-thought style reasoning~\cite{ammann2025question,wu2024gendec,wei2022chain}, but they are treated as removable knowledge units rather than privileged rationales.
They can be useful, redundant, or wrong; this noise is part of the dependency-estimation setting.

After all sources are collected, candidates are normalized, lightly de-duplicated, length-controlled, and capped.
During construction, we retain metadata indicating how each candidate was produced, such as whether it comes from task context, Wikipedia retrieval, generated subquestions, generated reasoning needs, or other automatically generated auxiliary candidates.
This metadata is used only for auditing and for designing balanced subset-sampling strategies.
It is not provided to \textbf{Knot} as a deployable input.
More broadly, reference answers, teacher outputs, construction-time scores, and construction-source identifiers are excluded from model inputs in order to reduce shortcut signals and keep the learned estimator focused on behavioral dependency rather than artifacts of the data construction process.

\subsection{Candidate-Space Audit}
\label{app:candidate_quality}

We audit the constructed candidate spaces to examine whether they provide a sufficiently diverse and answer-agnostic pool for dependency estimation.
The audit is intended as an analysis of the construction process rather than an evaluation of QA correctness.
Table~\ref{tab:candidate_quality_audit_appendix} reports candidate count, candidate length, audit-only answer-string coverage, and source composition on the test split.
The answer-coverage statistic is used only to characterize the candidate pool after construction.

\begin{table}[t]
\centering
\small
\setlength{\tabcolsep}{4.2pt}
\caption{
Candidate-space analysis on the test split.
\(|\mathcal{C}|\) denotes the number of candidate units per question.
Words denotes the average number of words per candidate unit.
Ans. cov. is an audit-only answer-string coverage statistic.
Ctx., Wiki, and Gen. denote the average number of task-context, Wikipedia-retrieval, and generated candidates, respectively, where generated candidates include generated subquestions and generated reasoning paths.
}
\label{tab:candidate_quality_audit_appendix}
\resizebox{\linewidth}{!}{%
\begin{tabular}{lrrrrrrrr}
\toprule
Dataset & Ex. & Avg. \(|\mathcal{C}|\) & Range & Words & Ans. cov. & Ctx. & Wiki & Gen. \\
\midrule
HotpotQA   & 500 & 11.99 & 11--12 & 23.31 & 83.8 & 2.99 & 5.00 & 4.00 \\
MedQA      & 500 & 9.00  & 9--9   & 24.16 & 24.4 & 0.00 & 5.00 & 4.00 \\
MMLU       & 500 & 11.73 & 5--12  & 24.32 & 39.0 & 0.00 & 5.00 & 6.73 \\
SQuAD      & 500 & 11.91 & 8--12  & 23.60 & 88.2 & 2.92 & 5.00 & 3.99 \\
TruthfulQA & 311 & 9.00  & 8--9   & 22.42 & 6.1  & 0.00 & 5.00 & 4.00 \\
\bottomrule
\end{tabular}
}
\end{table}

\subsection{Counterfactual Subset Sampling}
\label{app:subset_sampling}

For each question, teacher supervision is generated by removing sampled subsets
\(\mathcal{S}\subseteq\mathcal{C}(q)\) and comparing teacher behavior under the full and perturbed candidate spaces.
The subset sampler is deliberately designed rather than purely random.
This design reduces the risk that supervision quality is coupled to incidental patterns of candidate generation, which would make the resulting labels less informative for learning behavioral dependency.

The sampler includes singleton removals, representative pairs, high-relevance subsets, mixed-source subsets, complement-style removals, and low-signal control subsets.
Each sampled subset is later evaluated by the teacher model, whose answer under the perturbed condition is compared with the reference answer to produce the supervision signal.
Because this process requires additional teacher-model calls, we cap the number of generated subsets per question at 12 in the main supervision run.

\begin{figure}[t]
    \centering
    \includegraphics[width=\columnwidth]{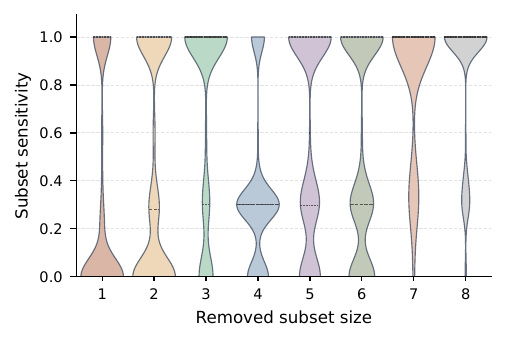}
    \caption{
    Retained subset-sensitivity labels by removed-subset size.
    This analysis checks whether larger removals tend to expose stronger behavioral dependency without assuming monotonicity for every individual question.
    }
    \label{fig:app_sensitivity_by_subset_size}
\end{figure}

\begin{table}[t]
\centering
\small
\setlength{\tabcolsep}{3.5pt}
\caption{
Major subset-sampling strategies used for counterfactual supervision on the training split.
Sens. is the mean final subset-sensitivity label after filtering.
}
\label{tab:subset_strategy_compact}
\resizebox{\linewidth}{!}{%
\begin{tabular}{lrrrr}
\toprule
Strategy & Generated & Kept & Keep \% & Sens. \\
\midrule
Random pair & 5{,}107 & 1{,}144 & 22.4 & 0.394 \\
Top discriminative & 5{,}046 & 3{,}925 & 77.8 & 0.659 \\
Large random & 5{,}032 & 2{,}868 & 57.0 & 0.404 \\
Top relevance & 5{,}013 & 3{,}573 & 71.3 & 0.642 \\
Mixed multi-source & 4{,}915 & 2{,}468 & 50.2 & 0.716 \\
Background-need control & 4{,}188 & 1{,}144 & 27.3 & 0.044 \\
Complement & 2{,}621 & 711 & 27.1 & 0.374 \\
\bottomrule
\end{tabular}
}
\end{table}

The final retained supervision rows are filtered to keep informative samples together with a controlled number of meaningful zero-sensitivity cases.
This filtering step improves the training signal but does not change the definition of subset sensitivity in Section~\ref{sec:method_supervision}.
Only examples that the teacher answers correctly under the full candidate context are retained, as judged by the hybrid scorer.
This ensures that sensitivity is measured from a clean full-context baseline.
Accordingly, the full-condition correctness in Table~\ref{tab:teacher_supervision_stats} is computed after filtering.
Table~\ref{tab:subset_strategy_compact} summarizes the retained subset-strategy families, and Fig.~\ref{fig:app_sensitivity_by_subset_size} checks how retained sensitivity labels vary with removed-subset size.

\subsection{Teacher Scoring and Hybrid Sensitivity Labels}
\label{app:teacher_scoring}

For each sampled subset, the teacher model is evaluated under three conditions:
\begin{equation}
\label{eq:app_teacher_conditions}
\begin{aligned}
\mathcal{K}_{\mathrm{full}} &= \mathcal{C}(q),\\
\mathcal{K}_{\mathrm{pert}} &= \mathcal{C}(q)\setminus\mathcal{S},\\
\mathcal{K}_{\emptyset} &= \emptyset .
\end{aligned}
\end{equation}
The full and perturbed conditions define the training target, while the no-knowledge condition is retained to assess whether the teacher appears to rely on supplied knowledge.
The no-knowledge condition is an evidence-ablation check rather than a random-guessing baseline: the prompt still preserves the task question and, for multiple-choice tasks, the answer space, but removes candidate knowledge and scores any abstention or unsupported output as incorrect.

Let \(y(q,\mathcal{K})\) be the teacher answer under condition \(\mathcal{K}\).
A task-aware canonicalizer maps raw outputs, option labels, and option texts into comparable answers.
Correctness is first scored with normalized exact match, option-space match, numeric match, safe containment, and high-overlap short-answer matching.
When the rule scorer is uncertain and judge mode is enabled, an LLM judge supplies a binary correctness judgment.
The final hybrid correctness score \(c(q,\mathcal{K})\in[0,1]\) uses the judge only for uncertain cases and otherwise keeps the rule-based score.

The subset sensitivity target combines non-negative correctness degradation and answer change:
\begin{equation}
\label{eq:app_hybrid_sensitivity}
\begin{aligned}
S(q,\mathcal{S})
=
\mathrm{clip}_{[0,1]}
\Big(
&\alpha\,
[c(q,\mathcal{K}_{\mathrm{full}})
 - c(q,\mathcal{K}_{\mathrm{pert}})]_{+} \\
&+ (1-\alpha)\,g(q,\mathcal{S})
\Big),
\end{aligned}
\end{equation}
where \(g(q,\mathcal{S})\) indicates task-aware answer change and \(\alpha=0.4\) in the main run.
For multiple-choice examples, answer equivalence is defined in the option space.
For generative examples, equivalence combines normalized matching and semantic-equivalence checks when available.

\begin{figure}[t]
    \centering
    \includegraphics[width=\columnwidth]{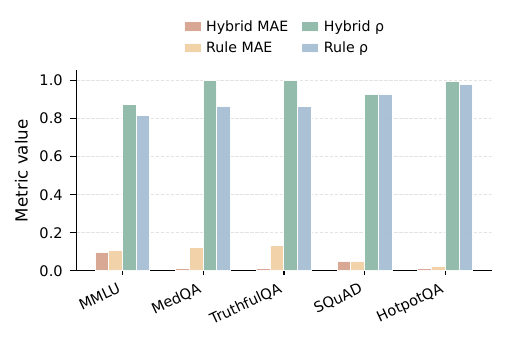}
\caption{
Reliability audit of sensitivity labels against forced LLM-judge sensitivity on a balanced audit subset.
Lower MAE and higher Spearman correlation indicate stronger agreement between rule-only or hybrid labels and the LLM-judge-derived reference signal.
}
    \label{fig:app_supervision_reliability_audit}
\end{figure}

\begin{figure}[t]
    \centering
    \includegraphics[width=\columnwidth]{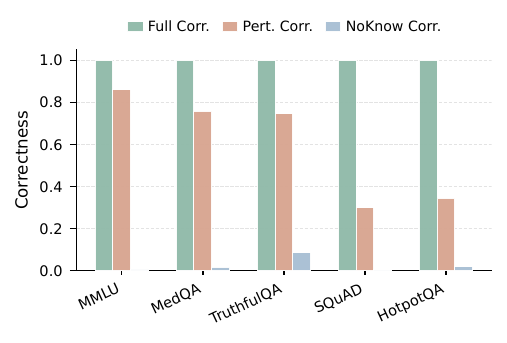}
    \caption{
    Teacher correctness under full, perturbed, and no-knowledge conditions on retained supervision rows.
    }
    \label{fig:app_teacher_correctness_conditions}
\end{figure}

Table~\ref{tab:teacher_supervision_stats} summarizes the retained supervision data.
Fig.~\ref{fig:app_supervision_reliability_audit} audits the hybrid labels against forced LLM-judge sensitivity on an audit subset.
Fig.~\ref{fig:app_teacher_correctness_conditions} further summarizes teacher correctness under full, perturbed, and no-knowledge conditions over the retained supervision rows.

\begin{table*}[t]
\centering
\small
\caption{Teacher-supervision statistics by split and dataset after retention filtering. Full Corr. and NoKnow Corr. denote teacher correctness under the full-candidate and no-knowledge conditions, respectively. Full Corr. is computed after filtering and is 1.000 by construction because rows with incorrect full-candidate teacher answers are excluded from the retained supervision data.}
\label{tab:teacher_supervision_stats}
\resizebox{\linewidth}{!}{%
\begin{tabular}{llrrrrrrrrr}
\toprule
Split & Dataset & Questions & Rows & $|\mathcal{C}|$ & Full Corr. & Pert. Corr. & NoKnow Corr. & Drop & AnsChange & Sensitivity \\
\midrule
train & MMLU & 416 & 4025 & 11.750 & 1.000 & 0.844 & 0.022 & 0.156 & 0.117 & 0.321 \\
dev & MMLU & 80 & 765 & 11.897 & 1.000 & 0.858 & 0.041 & 0.142 & 0.098 & 0.313 \\
test & MMLU & 433 & 4144 & 11.806 & 1.000 & 0.859 & 0.008 & 0.141 & 0.100 & 0.309 \\

train & MedQA & 395 & 1678 & 8.904 & 1.000 & 0.766 & 0.009 & 0.234 & 0.170 & 0.196 \\
dev & MedQA & 85 & 353 & 9.000 & 1.000 & 0.802 & 0.017 & 0.198 & 0.133 & 0.159 \\
test & MedQA & 424 & 1788 & 9.000 & 1.000 & 0.758 & 0.017 & 0.242 & 0.186 & 0.208 \\

train & TruthfulQA & 172 & 751 & 8.808 & 1.000 & 0.738 & 0.079 & 0.262 & 0.196 & 0.222 \\
dev & TruthfulQA & 38 & 148 & 9.000 & 1.000 & 0.824 & 0.095 & 0.176 & 0.135 & 0.151 \\
test & TruthfulQA & 182 & 792 & 8.996 & 1.000 & 0.746 & 0.086 & 0.254 & 0.192 & 0.217 \\
train & SQuAD & 410 & 4062 & 11.741 & 1.000 & 0.297 & 0.001 & 0.703 & 0.717 & 0.712 \\
dev & SQuAD & 84 & 844 & 11.910 & 1.000 & 0.296 & 0.000 & 0.704 & 0.712 & 0.709 \\
test & SQuAD & 406 & 4000 & 11.915 & 1.000 & 0.299 & 0.004 & 0.701 & 0.717 & 0.711 \\
train & HotpotQA & 396 & 3720 & 11.876 & 1.000 & 0.332 & 0.029 & 0.668 & 0.669 & 0.669 \\
dev & HotpotQA & 78 & 734 & 11.986 & 1.000 & 0.317 & 0.029 & 0.683 & 0.677 & 0.679 \\
test & HotpotQA & 382 & 3546 & 11.982 & 1.000 & 0.346 & 0.020 & 0.654 & 0.653 & 0.654 \\

\bottomrule
\end{tabular}
}
\end{table*}

\section{Knot Architecture and Training Details}
\label{app:kde_implementation}

\subsection{Deployable Inputs and Reproducible Configuration}
\label{app:model_config}

Table~\ref{tab:model_repro_config} summarizes the reproducible configuration of \textbf{Knot}.
The estimator operates on deployable information available at inference time, namely the question, candidate texts, optional raw task context, and online-computable lexical features.
Training uses the subset-level supervision described above, while inference only requires a forward pass over \((q,\mathcal{C}(q))\).

\begin{table}[t]
\centering
\small
\setlength{\tabcolsep}{4.5pt}
\caption{
Reproducible configuration of \textbf{Knot}.
The frozen encoder provides candidate representations, while all prediction heads and the set encoder are trained on subset-level supervision.
}
\label{tab:model_repro_config}
\resizebox{\linewidth}{!}{%
\begin{tabular}{ll}
\toprule
Component & Configuration \\
\midrule
Text encoder & \href{https://huggingface.co/Qwen/Qwen2.5-1.5B-Instruct-GGUF}{Qwen2.5-1.5B-Instruct}, frozen \\
Pooling & Mean pooling over token states \\
Max length & 256 tokens \\
Set encoder & 3-layer Transformer, no positional encoding \\
Latent factors & 30 \\
Lexical fusion & Enabled \\
Gate head & Enabled \\
Rank head & Enabled in main model \\
Optimizer & AdamW \\
Learning rate & \(1\times 10^{-4}\) \\
Weight decay & \(1\times 10^{-4}\) \\
Batch size & 16 question-grouped batches \\
Max epochs & 5 \\
Early stopping & Dev MAE / Spearman selection \\
\bottomrule
\end{tabular}
}
\end{table}

\subsection{Architecture Details}
\label{app:set_transformer_architecture}

For each candidate \(k_i\), we concatenate the question and candidate text and
encode the pair with a frozen text encoder:
\begin{equation}
\label{eq:app_candidate_encoding}
r_i = \mathrm{Pool}\big(\mathrm{Enc}_{\mathrm{frozen}}(q,k_i)\big),
\qquad
r_i\in\mathbb{R}^{d}.
\end{equation}
The encoder parameters are fixed during training, and only the lightweight
dependency-estimation modules are updated.

To contextualize candidates within the full candidate space, we apply a 3-layer
Transformer set encoder without positional encoding:
\begin{equation}
\label{eq:app_set_encoder}
h_{1:N}
=
\mathrm{SetEnc}(r_{1:N}).
\end{equation}
Since no positional encoding is used, the contextualization is
permutation-equivariant with respect to candidate order.
This allows the representation of each candidate to depend on other candidates
in \(\mathcal{C}(q)\), while avoiding an artificial dependence on the order in
which candidates are listed.

The lexical-fusion branch computes online features \(\ell_i\) from
the question and candidate pool, including candidate length, normalized
position, question--candidate lexical overlap, candidate-set size, and
redundancy statistics.
These features are projected and added to the contextual representation:
\begin{equation}
\label{eq:app_lexical_fusion}
\tilde{h}_i
=
h_i
+
\eta\,
\mathrm{MLP}_{\mathrm{lex}}(\ell_i).
\end{equation}

The main model then predicts three candidate-level quantities.
A gate head estimates a context-dependent usefulness scalar:
\begin{equation}
\label{eq:app_gate_head}
z_i=\sigma\big(w_z^\top \tilde{h}_i+b_z\big).
\end{equation}
A latent-factor head predicts the candidate's activation over \(R\) dependency
factors:
\begin{equation}
\label{eq:app_factor_activation}
\mathbf{a}_i
=
z_i\cdot
\sigma\big(W_a\tilde{h}_i+b_a\big),
\qquad
\mathbf{a}_i\in[0,1]^R.
\end{equation}
A rank head predicts a direct unit-ranking score:
\begin{equation}
\label{eq:app_rank_head}
s^{\mathrm{rank}}_i
=
\sigma\big(w_{\mathrm{rank}}^{\top}\tilde{h}_i+b_{\mathrm{rank}}\big).
\end{equation}
The subset-level prediction follows the latent-coverage aggregation defined in
Eq.~\ref{eq:method_noisy_or}.
For unit-level scoring, the model first computes the coverage-derived
singleton score
\(s^{\mathrm{cov}}_i=\widehat{S}_{\Theta}(q,\{k_i\};\mathcal{C}(q))\), and then mixes this score with the rank-head score:
\begin{equation}
\label{eq:app_unit_score_mix}
\widehat{u}_i
=
\rho s^{\mathrm{cov}}_i
+
(1-\rho)s^{\mathrm{rank}}_i,
\qquad
\rho=0.5 .
\end{equation}
Variants without the rank head set \(\widehat{u}_i=s^{\mathrm{cov}}_i\).

Training follows the objective in Eq.~\ref{eq:method_final_loss} while
the counterfactual reconstruction term \(\mathcal{L}_{\mathrm{cf}}\) follows
Eq.~\ref{eq:method_cf_loss}.
The unit-listwise term \(\mathcal{L}_{\mathrm{unit}}\) follows
Eq.~\ref{eq:method_unit_listwise_loss}.
Rows with fewer than two valid candidates, non-finite hints, non-positive hints,
or no hint contrast are skipped for this auxiliary term.
The unit-pair term \(\mathcal{L}_{\mathrm{pair}}\) follows
Eq.~\ref{eq:method_unit_pair_loss} and uses training-only weak pairs.

The entropy-sharpening term mildly discourages flat unit-score distributions:
\begin{equation}
\label{eq:app_unit_entropy_loss}
\mathcal{L}_{\mathrm{ent}}
=
\frac{1}{|\mathcal{Q}|}
\sum_{q\in\mathcal{Q}}
\frac{
H\!\left(\operatorname{softmax}(\widehat{\mathbf{u}}(q)/T_e)\right)
}{
\log |\mathcal{C}(q)|
}.
\end{equation}
Finally, the gate penalty is the mean candidate gate value:
\begin{equation}
\label{eq:app_gate_loss}
\mathcal{L}_{\mathrm{gate}}
=
\frac{1}{|\mathcal{Q}|}
\sum_{q\in\mathcal{Q}}
\frac{1}{|\mathcal{C}(q)|}
\sum_{k_i\in\mathcal{C}(q)}
z_i .
\end{equation}
In the main rank-aware configuration, \(T=T_e=0.10\), \(m=0.10\),
\(\lambda_{\mathrm{unit}}=0.20\), \(\lambda_{\mathrm{pair}}=0.20\),
\(\lambda_{\mathrm{ent}}=0.005\), and
\(\lambda_{\mathrm{gate}}=5\times10^{-5}\).

\subsection{Latent Coverage versus Additive Interaction}
\label{app:factor_vs_additive}

A direct way to model subset sensitivity is to use an additive set function, where the sensitivity of a removed subset is decomposed into first-order unit effects and explicit pairwise interactions.
We implement this additive structured approximation as the \textbf{w/o latent coverage} ablation.
In that variant, each candidate receives a first-order score \(u_i\), each selected pair receives a learned interaction \(I_{ij}\), and subset sensitivity is predicted as
\begin{equation}
\label{eq:app_additive_surrogate}
\begin{aligned}
\widehat{S}_{\mathrm{add}}(q,\mathcal{S})
=
\mathrm{clip}_{[0,1]}
\Big(
&\sum_{i\in\mathcal{S}} u_i \\
&+
\sum_{\substack{i<j\\i,j\in\mathcal{S}}} I_{ij}
\Big).
\end{aligned}
\end{equation}
The pair interaction is computed from contextualized candidate representations using
\([h_i,h_j,|h_i-h_j|,h_i\odot h_j]\), passed through a small MLP, symmetrized, and gated by \(\sqrt{z_i z_j}\).

This additive form gives the ablation access to explicit second-order interactions, but it is less suitable for noisy candidate spaces with redundant or substitutable evidence.
When two candidates express the same dependency, their additive scores can over-count the same behavioral factor.
When candidates are substitutes, a linear sum does not naturally saturate.
\textbf{Knot} represents each candidate as covering latent dependency factors and aggregates removed subsets with the noisy-OR coverage function in Eq.~\ref{eq:method_noisy_or}.
This aggregation saturates repeated evidence for the same factor while still accumulating evidence across distinct factors.
The architectural ablation in Table~\ref{tab:rq3_architectural_ablation} tests this modeling choice directly.

\subsection{Amortized Estimation Benefit}
\label{app:amortized_benefit}

Perturbation-based unit attribution requires new QA-model calls whenever a candidate space changes.
LOO requires one removal call per candidate, and MC-Shapley requires many prefix-removal calls.
\textbf{Knot} moves this cost to offline supervision construction: after training, it estimates subset and unit dependency from \((q,\mathcal{C}(q))\) without additional QA-model perturbation.

\section{Experimental Details and Additional Results}
\label{app:additional_results}

\subsection{Model Backend and Baseline Cost Accounting}
\label{app:baseline_runtime_details}

The main experiments are conducted on one NVIDIA A100 80GB GPU.
We use the open-weight backend \texttt{Qwen/Qwen3.5-35B-A3B} as both the target QA model and the teacher model.
This backend provides a capable and controllable QA model for large-scale perturbation experiments while keeping the setup reproducible in a controlled environment.
Due to the cost of live perturbation and LLM-judge calls, the main study evaluates one strong base model.
Since our formulation treats the QA model as a black box, the same supervision and amortized-estimation protocol can be applied to other LLM backends.

The full-test comparison focuses on deployable methods: Size, BM25, Lex-Ridge, Lex-HGB, Lex-ET, Lex-MLP, and \textbf{Knot}, all of which avoid live QA perturbation at inference time.
The sampled expensive panel additionally includes LLM-Rank, LOO, MC-S4, and MC-S16.
These methods provide useful behavioral reference points, but their request cost grows with the candidate-space size or perturbation budget.
For cost reporting, Calls/Q denotes cache-independent backend requests per question, so cached reruns do not reduce the reported inference-time cost.
Table~\ref{tab:sampled_expensive_runtime_details} reports the cost and evaluation metrics for the sampled expensive-method panel.

\begin{table}[t]
\centering
\small
\setlength{\tabcolsep}{4.2pt}
\caption{Results on the sampled expensive-evaluation set with the R=30, L=3 Knot checkpoint. Calls/Q reports cache-independent backend requests per question required for dependency estimation, assuming \(N=12\) candidate units.}
\label{tab:sampled_expensive_runtime_details}
\resizebox{\linewidth}{!}{%
\begin{tabular}{lrrrrcc}
\toprule
Method & MAE & RMSE & Pearson & Spearman & Calls/Q & Cost \\
\midrule
LLM-Rank & 0.399 & 0.562 & 0.408 & 0.479 & 1 & \(1\) \\
LOO & 0.394 & 0.560 & 0.236 & 0.262 & 13 & \(1+N\) \\
MC-S4 & 0.273 & 0.390 & 0.559 & 0.569 & 49 & \(1+4N\) \\
MC-S16 & 0.272 & 0.383 & 0.570 & 0.577 & 193 & \(1+16N\) \\
\textbf{Knot} & \textbf{0.206} & \textbf{0.303} & \textbf{0.731} & \textbf{0.715} & \textbf{0} & \(\mathbf{0}\) \\
\bottomrule
\end{tabular}
}
\end{table}

\subsection{Architectural Hyperparameter Ablations}
\label{app:arch_hyperparam_ablation}

We further evaluate Knot on the test split by varying two key architectural hyperparameters:
the number of latent dependency factors \(R\) and the number of candidate-set encoder layers \(L\). These two parameters control different aspects of the model capacity:
\(R\) determines the dimensionality of the latent coverage bottleneck used to represent
heterogeneous dependency patterns, while \(L\) controls how strongly each candidate
representation is contextualized by the other candidates in \(\mathcal{C}(q)\).

\begin{figure*}[t]
    \centering
    \begin{subfigure}[t]{0.48\textwidth}
        \centering
        \includegraphics[width=\linewidth]{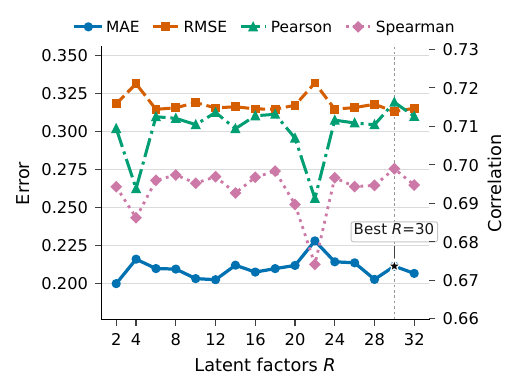}
        \caption{Latent factor capacity.}
        \label{fig:knot_factor_capacity}
    \end{subfigure}
    \hfill
    \begin{subfigure}[t]{0.48\textwidth}
        \centering
        \includegraphics[width=\linewidth]{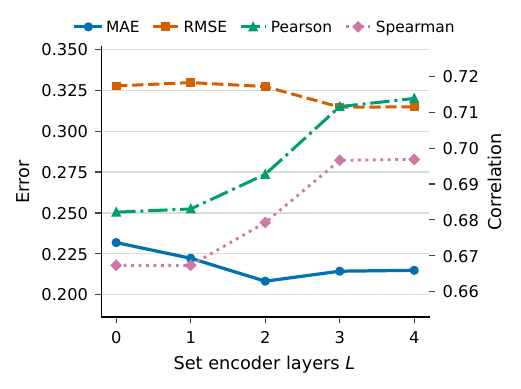}
        \caption{Set encoder depth.}
        \label{fig:knot_set_depth}
    \end{subfigure}
    \caption{
    Architectural hyperparameter ablations of Knot on the full test split.
    Left: varying the number of latent dependency factors \(R\).
    Right: varying the number of set encoder layers \(L\).
    The marked point in the latent-factor plot indicates the selected default capacity,
    chosen by considering error, correlation, and ranking metrics jointly.
    }
    \label{fig:knot_arch_hyperparam_ablation}
\end{figure*}

Fig.~\ref{fig:knot_arch_hyperparam_ablation} shows that Knot is relatively stable
across a broad range of latent factor capacities and set encoder depths. For latent
capacity, very small factor spaces already achieve competitive MAE, suggesting that a
compact bottleneck can capture a large portion of the subset-level sensitivity signal.
However, \(R=30\) provides the strongest overall trade-off: it obtains the lowest RMSE
and the best Pearson and Spearman correlations among the tested settings. Although
smaller values such as \(R=2\) can slightly reduce MAE, they are weaker on the
correlation and ranking metrics. Since our goal is to minimize pointwise
prediction error while preserving the relative ordering of sensitive subsets, we
select \(R=30\) as the default latent factor capacity.

The set encoder depth ablation in Fig.~\ref{fig:knot_arch_hyperparam_ablation} further shows
that candidate-set contextualization is important. Removing the set encoder
(\(L=0\)) consistently degrades performance, indicating that dependency estimation
benefits from modeling interactions among candidates rather than scoring each candidate
independently. Increasing the depth to \(L=3\) yields clear improvements in both error
and correlation metrics. While \(L=4\) gives a marginally higher Pearson score and a
similar Spearman score, it does not improve RMSE and introduces additional depth. We
therefore use \(L=3\) as the default setting, as it provides a strong accuracy--complexity
trade-off and captures most of the benefit of set-level contextualization.

Overall, these ablations support the default architecture used in the main experiments:
\(R=30\) latent factors and \(L=3\) set encoder layers.

\subsection{Optimization Variants} \label{app:optimization_variants}

The main experiments use \textbf{Knot} as the final model configuration.
This section isolates how different unit-ranking objectives affect subset-level calibration and unit-level faithfulness.
All variants instantiate the objective in Eq.~\ref{eq:method_final_loss}, and keep the candidate spaces, supervision data, and deployable inputs fixed.
They differ only in the unit-ranking objective weights or in whether a separate rank head is used.

The comparison starts from the lexical-fusion base model without the rank-aware objective.
This model already reconstructs subset sensitivity with \(\mathcal{L}_{\mathrm{cf}}\) and uses the weak listwise unit loss \(\mathcal{L}_{\mathrm{unit}}\), but its unit score is still the singleton score induced by the latent coverage pathway.
It therefore tests whether subset reconstruction plus weak distribution matching is sufficient to produce useful unit rankings.
This setting uses \(\lambda_{\mathrm{unit}}=0.10\), \(T=0.20\), \(\lambda_{\mathrm{gate}}=5\times10^{-5}\), and sets \(\lambda_{\mathrm{pair}}=\lambda_{\mathrm{ent}}=0\).

The \emph{unit-strong} variant asks whether stronger listwise unit supervision alone improves local ranking.
It keeps the same scoring architecture as the lexical-fusion base model, but sharpens the weak unit-hint distribution and increases the influence of \(\mathcal{L}_{\mathrm{unit}}\) during training.
Concretely, it uses \(\lambda_{\mathrm{unit}}=0.30\), \(T=0.10\), \(\lambda_{\mathrm{gate}}=5\times10^{-5}\), and keeps \(\lambda_{\mathrm{pair}}=\lambda_{\mathrm{ent}}=0\).
The \emph{pair-rank} variant adds direct pairwise ordering constraints between candidates whose weak hints differ, separating the question of whether the top candidates are ordered correctly from the softer distribution-matching objective in \(\mathcal{L}_{\mathrm{unit}}\).
It uses \(\mathcal{L}_{\mathrm{pair}}\) with \(\lambda_{\mathrm{pair}}=0.20\) and margin \(m=0.10\), while still using \(\lambda_{\mathrm{unit}}=0.20\) and \(T=0.10\).
This variant does not add a separate rank head and sets \(\lambda_{\mathrm{ent}}=0\).
The \emph{w/o unit-listwise} variant removes \(\mathcal{L}_{\mathrm{unit}}\), testing whether subset-level reconstruction alone can recover useful unit ordering or whether subset-derived unit hints are needed to make singleton scores informative.

The final \textbf{Knot} configuration combines the useful pieces from these analyses.
It keeps subset reconstruction as the primary signal, adds both \(\mathcal{L}_{\mathrm{unit}}\) and \(\mathcal{L}_{\mathrm{pair}}\), and introduces the rank head in Eq.~\ref{eq:app_rank_head} so that unit ordering is not forced to be expressed only through the subset-coverage pathway.
The final unit score mixes coverage and rank-head scores with \(\rho=0.5\), and a small entropy term \(\mathcal{L}_{\mathrm{ent}}\) discourages an overly flat top-unit distribution.
The main setting uses \(\lambda_{\mathrm{unit}}=0.20\), \(\lambda_{\mathrm{pair}}=0.20\), \(\lambda_{\mathrm{ent}}=0.005\), and \(\lambda_{\mathrm{gate}}=5\times10^{-5}\).

Table~\ref{tab:appendix_training_objective_variants} reports held-out subset prediction and unit-ranking quality for these training-objective variants.
The lexical-fusion base is already competitive, with test MAE \(0.212\) and Spearman \(0.685\).
Increasing the emphasis on unit-oriented objectives does not automatically improve subset calibration: the unit-strong and pair-rank variants trail the base on the held-out subset metrics.
Removing the unit-listwise term gives the best subset-only result in this diagnostic group, with MAE \(0.203\), Pearson \(0.710\), and Spearman \(0.696\).
By contrast, the pair-rank objective gives the strongest unit-ranking scores, reaching NDCG@1 \(0.810\) and NDCG@3 \(0.822\).
This trade-off shows why we treat the objective variants as diagnostics rather than selecting the final model by subset error alone.
The final \textbf{Knot} configuration balances subset reconstruction with rank-aware unit supervision, remaining competitive on subset prediction while preserving faithful unit-level behavioral estimates.

\begin{table}[t]
\centering
\small
\setlength{\tabcolsep}{2.4pt}
\caption{
Additional training-objective variants for \textbf{Knot}.
The lexical-fusion base removes the rank-aware unit objective; the remaining variants use the same architecture family and supervision data, but modify the auxiliary unit-ranking objective.
All metrics are computed on the held-out test split.
Higher is better except for MAE.
}
\label{tab:appendix_training_objective_variants}
\resizebox{\linewidth}{!}{%
\begin{tabular}{@{}lrrrrr@{}}
\toprule
Variant & MAE & Pear. & Sp. & NDCG@1 & NDCG@3 \\
\midrule
Lexical-fusion base
& 0.212 & 0.703 & 0.685 & 0.792 & 0.807 \\
Unit-strong
& 0.242 & 0.685 & 0.662 & 0.804 & 0.798 \\
Pair-rank
& 0.235 & 0.690 & 0.664 & \textbf{0.810} & \textbf{0.822} \\
w/o unit-listwise
& \textbf{0.203} & \textbf{0.710} & \textbf{0.696} & 0.796 & 0.810 \\
\bottomrule
\end{tabular}
}
\end{table}

\subsection{Dependency-Based Reliability Details}
\label{app:reliability_details}

We evaluate whether knowledge-dependency estimates can help identify unreliable answers.
At deployment time, a dependency estimator can directly rank candidate units by its predicted unit scores.
Here, however, reliability assessment is conducted as an additional behavior-audit experiment: for each method, we first select the top-\(k\) candidate units according to its unit scores, then measure how the QA model behaves when these units are removed or retained alone.
Let \(\mathrm{Drop@}k\) denote the answer degradation after removing the top-\(k\) units, and let \(\mathrm{Suff@}k\) denote answer quality when only the top-\(k\) units are retained.
We define the audited question-level risk score as
\begin{equation}
\label{eq:app_dependency_risk}
\begin{aligned}
\mathrm{Risk@}k
=
\mathrm{clip}_{[0,1]}
\Big(
&\frac{1}{2}\mathrm{Drop@}k \\
&+
\frac{1}{2}\bigl(1-\mathrm{Suff@}k\bigr)
\Big).
\end{aligned}
\end{equation}
A high score indicates that the answer is sensitive to removing the selected units, while those units alone are insufficient to stably preserve the answer.

Questions are sorted by \(\mathrm{Risk@}k\), and we evaluate whether full-context answer errors concentrate near the top of the ranked list.
We report AUROC, AUPRC, the error rate among the top-risk fraction, and lift over the overall error rate, using a top-risk fraction of \(10\%\), as shown in Table~\ref{tab:appendix_risk_indication}.
For LLM-based ranking, the valid set can differ because examples are excluded when the LLM fails to follow the required output format within three attempts, so its overall error rate is not always directly comparable to methods evaluated on the full valid audit set.

Overall, the results show that high-risk questions identified by dependency-based scores are substantially enriched for full-answer errors:
the top \(10\%\) risk slice consistently exceeds the overall error rate, and Knot achieves the strongest Risk@2 concentration, with Err.@10\% \(=0.700\) and a \(3.33\times\) lift despite requiring no test-time perturbation or LLM ranking calls.

\begin{table}[t]
\centering
\small
\setlength{\tabcolsep}{4.0pt}
\caption{
Dependency-based risk indication on the sampled test set.
Risk@1 and Risk@2 are evaluation-time audited risk scores computed from the behavioral consequences of the top-1 and top-2 predicted knowledge units, respectively.
Err. denotes the full-answer error rate on the valid examples for each method, and is computed by scoring fresh full-context QA outputs with the task-aware hybrid scoring procedure described in Appendix~\ref{app:teacher_scoring}.
Err.@10\% reports the error rate among the top 10\% highest-risk questions, and Lift is Err.@10\% divided by Err.
}
\label{tab:appendix_risk_indication}
\resizebox{\linewidth}{!}{%
\begin{tabular}{lrrrrr}
\toprule
Method & AUROC & AUPRC & Err. & Err.@10\% & Lift \\
\midrule
\multicolumn{6}{l}{\textit{Risk@1}} \\
\midrule
Lex-HGB  & 0.710 & 0.470 & 0.210 & 0.500 & 2.381 \\
Lex-ET   & 0.715 & 0.497 & 0.210 & 0.500 & 2.381 \\
BM25     & 0.693 & 0.379 & 0.210 & 0.500 & 2.381 \\
LLM-Rank & 0.764 & 0.542 & 0.211 & 0.667 & 3.158 \\
LOO      & 0.775 & 0.488 & 0.210 & 0.600 & 2.857 \\
MC-S4    & 0.790 & 0.556 & 0.210 & 0.600 & 2.857 \\
MC-S16   & \textbf{0.841} & \textbf{0.640} & 0.210 & \textbf{0.700} & \textbf{3.333} \\
\textbf{Knot} & 0.699 & 0.411 & 0.210 & 0.600 & 2.857 \\
\midrule
\multicolumn{6}{l}{\textit{Risk@2}} \\
\midrule
Lex-HGB  & 0.717 & 0.475 & 0.210 & 0.600 & 2.857 \\
Lex-ET   & 0.694 & 0.493 & 0.210 & 0.500 & 2.381 \\
BM25     & 0.587 & 0.274 & 0.210 & 0.300 & 1.429 \\
LLM-Rank & 0.765 & 0.427 & 0.211 & 0.444 & 2.105 \\
LOO      & 0.734 & 0.416 & 0.210 & 0.500 & 2.381 \\
MC-S4    & \textbf{0.769} & 0.487 & 0.210 & 0.600 & 2.857 \\
MC-S16   & 0.717 & 0.479 & 0.210 & 0.500 & 2.381 \\
\textbf{Knot} & 0.677 & \textbf{0.497} & 0.210 & \textbf{0.700} & \textbf{3.333} \\
\bottomrule
\end{tabular}
}
\end{table}

\section{Extended Related Work}
\label{app:extended_related_work}

\subsection{Knowledge Attribution and Faithful Explanation}

Explainable QA and LLM reasoning research has studied rationale extraction, supporting evidence identification, and reasoning decomposition.
Faithful explanation benchmarks and rationale learning methods emphasize explanations that reflect model behavior rather than plausible post-hoc justifications~\cite{deyoung2020eraser,jacovi2020towards}.
In multi-hop QA and retrieval-augmented generation, supporting evidence selection is often used to make reasoning processes more interpretable~\cite{lewis2020retrieval,izacard2023atlas}.
Chain-of-thought prompting and self-consistency further show that intermediate reasoning traces can improve reasoning performance and provide useful explanatory structure~\cite{wei2022chain,wang2023selfconsistency}.

These lines of work typically rely on relatively clean rationales, supporting passages, or semantically meaningful reasoning traces.
In contrast, Knot studies noisy and non-unique candidate knowledge spaces, where candidate units may be redundant, partially relevant, incomplete, or structurally ambiguous.
Our goal is not to recover a gold rationale, but to estimate the behavioral dependency of model predictions on candidate knowledge units.

\subsection{Perturbation-based and Counterfactual Attribution}

Perturbation-based attribution methods estimate importance by measuring prediction changes under input modification.
Classical methods such as SHAP~\cite{lundberg2017unified} and representation erasure~\cite{li2016understanding} have inspired many perturbation-based explanation methods in NLP.
Recent work further studies counterfactual and intervention-based explanations through input editing, causal mediation, and reasoning interventions~\cite{vig2020investigating,meng2022locating,kaushik2020learning}.

Although perturbation methods provide direct behavioral evidence, they often require explicit intervention at inference time and usually operate at the token, feature, or representation level.
They also become expensive and unstable when candidate knowledge units are redundant, overlapping, or substitutable.
Knot learns an amortized estimator from subset-level counterfactual supervision, enabling perturbation-free inference-time estimation over structured knowledge units.

\subsection{Uncertainty and Reliability Estimation}

Uncertainty estimation is central to reliable large language models.
Recent work studies hallucination detection and uncertainty estimation through output consistency, semantic agreement, and calibration analysis.
SelfCheckGPT~\cite{manakul2023selfcheckgpt} estimates hallucination via sampling consistency, while semantic entropy and related methods cluster semantically equivalent generations to estimate uncertainty at the meaning level~\cite{farquhar2024detecting,kossen2024semantic,kuhn2023semantic,tong2026semantic}.
Other work investigates confidence calibration and uncertainty-aware model behavior~\cite{kadavath2022language,lin2022teaching}.

These approaches primarily estimate reliability from output variability, confidence signals, or semantic dispersion among generated answers.
Knot provides a complementary view by modeling how predictions depend on candidate knowledge units.
This dependency-centered perspective yields reliability signals grounded in behavioral sensitivity to supporting knowledge.

\subsection{Structured and Amortized Attribution}

Our work is also related to amortized inference and structured attribution modeling.
Amortized inference methods learn reusable estimators that approximate expensive inference or optimization procedures~\cite{gershman2014amortized,marino2018iterative}.
Attribution research studies feature interactions, sparse explanation structure, and subset-level contribution modeling~\cite{tsang2018detecting,frye2020asymmetric,covert2021explaining}.
Cooperative game-theoretic methods, including Shapley-value attribution, analyze how feature subsets jointly contribute to predictions~\cite{lundberg2017unified}.

Different from prior feature-level attribution methods, we formulate knowledge dependency estimation as a structured amortized attribution problem over noisy candidate knowledge spaces.
Knot learns reusable unit-level sensitivity estimators from subset-level counterfactual supervision and uses a redundancy-aware latent coverage structure to handle overlapping and substitutable knowledge units.

\end{document}